%% file: arxiv.tex
\documentclass{article}

\usepackage[accepted]{icml2020}

\PassOptionsToPackage{numbers,compress}{natbib}

\usepackage{microtype}
\usepackage{graphicx}

\usepackage{subcaption}

\usepackage{booktabs}
\usepackage{verbatim}

\usepackage{hyperref}

\usepackage{algorithm}
\usepackage[noend]{algpseudocode}

\usepackage{times}
\usepackage{latexsym}

\usepackage{url}

\usepackage[xindy]{glossaries}
\usepackage[capitalise]{cleveref}
\usepackage{multicol}

\input{math_commands}

\input{packages}
\input{acronyms}

\input{macros}

\usepackage{varwidth}

\usepackage{paralist}
\usepackage{mathtools}

\usepackage[T1]{fontenc}
\usepackage[font=small,labelfont=bf,tableposition=top]{caption}

\DeclareCaptionLabelFormat{andtable}{#1~#2  \&  \tablename~\thetable}

\floatname{algorithm}{Algorithm}

\renewcommand{\paragraph}[1]{\textbf{#1}\ }

\DeclareMathOperator{\select}{\module{select}}

\DeclareMathOperator{\attn}{\boldsymbol{\alpha}}

\DeclareMathOperator{\encoder}{enc}

\algrenewcommand\algorithmicindent{0.5em}%

\makeatletter
\newcommand*{\skipnumber}[2][1]{%
   {\renewcommand*{\alglinenumber}[1]{}\State #2}%
   \addtocounter{ALG@line}{-#1}}
\makeatother

\begin{document}

\twocolumn[
    \icmltitle{Learning Reasoning Strategies in End-to-End Differentiable Proving}

    \icmlsetsymbol{equal}{*}
    
    \begin{icmlauthorlist}
    \icmlauthor{Pasquale Minervini}{ucl}
    \icmlauthor{Sebastian Riedel}{ucl,fair}
    \icmlauthor{Pontus Stenetorp}{ucl}
    \icmlauthor{Edward Grefenstette}{ucl,fair}
    \icmlauthor{Tim Rocktäschel}{ucl,fair}
    \end{icmlauthorlist}
    
    \icmlaffiliation{ucl}{UCL Centre for Artificial Intelligence, University College London}
    \icmlaffiliation{fair}{Facebook AI Research}
    
    \icmlcorrespondingauthor{Pasquale Minervini}{p.minervini@ucl.ac.uk}
    
    \icmlkeywords{Neuro-Symbolic Reasoning,Representation Learning,Relational Learning}
    
    \vskip 0.3in
]

\printAffiliationsAndNotice

\begin{abstract}
Attempts to render deep learning models interpretable, data-efficient, and robust have seen some success through hybridisation with rule-based systems, for example, in~\glspl{NTP}.
These neuro-symbolic models can induce interpretable rules and learn representations from data via back-propagation, while providing logical explanations for their predictions.
However, they are restricted by their computational complexity, as they need to consider all possible proof paths for explaining a goal, thus rendering them unfit for large-scale applications.
We present \glspl{CTP}, an extension to \glspl{NTP} that learns an optimal rule selection strategy via gradient-based optimisation.
We show that \glspl{CTP} are scalable and yield state-of-the-art results on the \glsentryshort{CLUTRR} dataset, which tests \emph{systematic generalisation} of neural models by learning to reason over smaller graphs and evaluating on larger ones.
Finally, \glspl{CTP} show better link prediction results on standard benchmarks in comparison with other neural-symbolic models, while being explainable.
All source code and datasets are available online.~\footnote{At \url{https://github.com/uclnlp/ctp}}
\end{abstract}

\section{Introduction}
Neural \gls{NLU} systems---wherein a deep neural network is used as a function approximator~\citep{DBLP:journals/nature/LeCunBH15,DBLP:books/daglib/0040158}---have been extremely successful at various natural language tasks, such as \gls{QA} and \gls{NLI}~\citep{DBLP:series/synthesis/2017Goldberg}, achieving strong generalisation results on datasets available for these tasks~\citep{DBLP:conf/iclr/SeoKFH17,DBLP:conf/ijcai/HuPHQW018,DBLP:journals/corr/ShenHGC16,DBLP:conf/iclr/HuangZSC18}.
Even strong performance on NLU problems have been recently achieved with advent of large models pre-trained via self-supervision, such as BERT~\citep{DBLP:conf/naacl/DevlinCLT19}, XLNet~\citep{DBLP:journals/corr/abs-1906-08237}, and RoBERTa~\citep{DBLP:journals/corr/abs-1907-11692}.
\paragraph{Generalisation in Neural Models}
However, there are growing concerns about the ability of \gls{NLU} systems, and neural networks more generally, to generalise in a systematic and robust way~\citep{DBLP:conf/iclr/BahdanauMNNVC19,DBLP:conf/icml/LakeB18,DBLP:conf/cvpr/JohnsonHMFZG17,DBLP:conf/emnlp/SinhaSDPH19}.
For instance, \citet{DBLP:conf/emnlp/JiaL17} highlight the brittleness of \gls{NLU} systems to adversarial examples, while \citet{DBLP:conf/naacl/GururanganSLSBS18,DBLP:conf/emnlp/KaushikL18} show that neural \gls{NLU} models tend to exploit annotation artefacts and spurious correlations in the data.
Furthermore, analysing and supervising the inner workings of such models is not trivial, due to their inherent black-box nature~\citep{kim,DBLP:journals/cacm/Lipton18}.
More generally, \citet{GARNELO201917} emphasise several limitations of neural models, in terms of
\begin{inparaenum}[\itshape i\upshape)]
\item data inefficiency and high sample complexity---the need of high volumes of training data in order to be effective,
\item poor generalisation---modern neural models may not produce the correct predictions when exposed to data outside the training distribution, and
\item lack of interpretability---such models are \emph{black boxes} where internal representations and computations are hardly interpretable by humans.
\end{inparaenum}
In this vein, \citet{DBLP:conf/emnlp/SinhaSDPH19} measured and compared the systematic generalisation abilities of several neural models (including very strong baselines such as BERT~\citep{DBLP:conf/naacl/DevlinCLT19} and \glspl{GAT}~\citep{DBLP:conf/iclr/VelickovicCCRLB18}) on the task of answering questions about family relationship graphs, by evaluating on held-out combinations of reasoning patterns and by adding curated distracting noisy facts.
Interestingly, they found that performance degrades monotonically for every model in their pool as they increase the complexity of the relational graph, highlighting the challenge of systematic generalisation~\citep{DBLP:conf/icml/LakeB18,DBLP:journals/corr/abs-1811-07017}.
\paragraph{Neuro-Symbolic Reasoning}
A promising direction for overcoming these issues consists in combining \emph{neural models} and \emph{symbolic reasoning} given their complementary strengths and weaknesses~\citep{DBLP:conf/aaaiss/GarcezBRFHIKLMS15,evans2018learning,GARNELO201917}.
We focus on \glspl{NTP}~\citep{DBLP:conf/nips/Rocktaschel017}, a family of neuro-symbolic reasoning models: \glspl{NTP} are continuous relaxations of the backward-chaining reasoning algorithm that replace discrete symbols with their continuous embedding representations.
\glspl{NTP} have interesting properties: they can jointly learn representations and interpretable rules from data via backpropagation, and can potentially combine such rules in ways that may have not been observed during training.
However, a major limitation in \glspl{NTP} is that, during training, they need to consider \emph{all rules} for explaining a given goal or sub-goal.
This quickly renders them ineffective in settings requiring a large number of rules or reasoning steps.
\paragraph{Conditional Theorem Provers}
For addressing limitations of \glspl{NTP}, we propose \glspl{CTP}, an extension that is able to learn an adaptive strategy for selecting subsets of rules to consider at each step of the reasoning process.
This is achieved by a $\module{select}$ module that, given a goal, produce the rules needed for proving it.
Predicates and constants in the produced rules lie in a continuous embedding space. Hence, the $\module{select}$ module is end-to-end differentiable, and can be trained jointly with the other modules via gradient-based optimisation.
\section{End-to-End Differentiable Proving} \label{sec:ntp}
\begin{algorithm*}[t]
    \caption{Overview of the neural backward chaining algorithm proposed by \citet{DBLP:conf/nips/Rocktaschel017} -- intuitively, it recursively proves each goal with all rules in the KB (OR module) and, for each rule, it proves its premise (AND module), up to $d$ recursion steps.} \label{alg:bc}
    \vspace{-10pt}
    \begin{multicols}{2}
    \begin{algorithmic}[1]
    \Function{$\module{or}$}{$G$, $d$, $\state$} \label{ln:or}
        \For{$\lst{H} \lif \lss{B} \in \kb$} \label{ln:all}
            \For{$\state \in \module{and}\left(\lss{B}, d, \module{unify}(\lst{H}, \lst{G}, \state)\right)$}
                \State {\bf yield} $\state$
            \EndFor
        \EndFor
    \EndFunction
    \end{algorithmic}
    \begin{algorithmic}[1]
    \Function{$\module{and}$}{$\lss{B}$, $d$, $S$} \label{ln:and}
        \State{\algorithmicif\ $\lss{B} = []$ or $d = 0$\ \algorithmicthen\ {\bf yield} $\state$ \algorithmicelse}
            \For{$\state' \in \module{or}\left(\module{sub}(\lss{B}_{0}, \state_\subs), d - 1, \state\right)$}
                \For{$\state'' \in \module{and}(\lss{B}_{1:}, d, \state')$}
                    \State {\bf yield} $\state''$
                \EndFor
            \EndFor
    \EndFunction
    \end{algorithmic}
    \begin{algorithmic}[1]
    \Function{$\module{unify}$}{$\ls{H}$, $\ls{G}$, $S = (\state_\subs, \state_\success)$} \label{ln:unify}
        \State $\state'_\subs = \state_\subs \bigcup_i T_{i}$
        \skipnumber[2] \State with $T_{i} = \begin{dcases}
                \{H_i/G_i\} & \text{if } H_i \in \set{V}\\
                \{G_i/H_i\} & \text{if } G_i \in \set{V}, H_i \not\in \set{V}\\
                \emptyset & \text{otherwise}
            \end{dcases}$
        \State $\state'_\success = \min \left\{ \state_\success \right\} \bigcup_{H_i, G_i \not\in \set{V}} \left\{ \krnl(\params_{H_i}, \params_{G_i}) \right\}$
        \State \Return $(\state'_\subs, \state'_\success)$
    \EndFunction
    \end{algorithmic}
    \end{multicols}
    \vspace{-10pt}
\end{algorithm*}

\glspl{NTP}~\citep{DBLP:conf/nips/Rocktaschel017} are a continuous relaxation of the \emph{backward chaining algorithm}~\citep{DBLP:books/daglib/0023820}: this algorithm works backward from the goal, chaining through rules to find known facts supporting the proof.
Given a \emph{query} (or \emph{goal}) $G$, backward chaining first attempts to unify it with the facts available in a given \gls{KB}.
If no matching fact is available, it considers all rules $\lst{H} \lif \lst{B}$, where $\lst{H}$ denotes the \emph{head} (or \emph{consequence}) and $\lst{B}$ the \emph{body} (or \emph{premise}), and $\lst{H}$ can be unified with the query $G$ resulting in a substitution for the variables contained in $\lst{H}$.
Then, the backward chaining algorithm applies the substitution to the body $\lst{B}$, and recursively attempts to prove the atoms contained therein.
Backward chaining can be seen as a type of $\module{and}$/$\module{or}$ search: $\module{or}$ because the goal can be proven by any rule in the \gls{KB}, and $\module{and}$ because all the conjuncts in the premise of a rule must be proven.
\begin{example}[Backward Chaining] \label{ex:1}
Consider a \gls{KB} composed by the facts $\rel{p}(\const{rick}, \const{beth})$ and $\rel{p}(\const{beth}, \const{morty})$, and by the rule $\rel{g}(\var{X}, \var{Y}) \lif \rel{p}(\var{X}, \var{Z}), \rel{p}(\var{Z}, \var{Y})$, where $\rel{p}$ and $\rel{g}$ denote the relationships \emph{parent} and \emph{grandparent}, respectively.
The goal $G = \rel{g}(\const{rick}, \const{morty})$ can be proven by unifying $G$ with the head of the rule $\rel{g}(\var{X}, \var{Y})$, with the substitution $\{ \var{X}/\const{rick}, \var{Y}/\const{morty} \}$, and then by recursively proving the subgoals $\rel{p}(\const{rick}, \var{Z}), \rel{p}(\var{Z}, \const{morty})$, which hold true for the substitution $\{ \var{Z}/\const{beth} \}$. $\triangle$
\end{example}
\glspl{NTP} make this reasoning process more flexible and end-to-end differentiable by replacing the comparison between symbols with a soft matching of their respective trainable dense vector representations.
\glspl{NTP} recursively build a neural network enumerating all possible proof paths for proving a query given \gls{KB}, and aggregate all proof scores via max pooling.
They do so by relying on three modules: a \emph{unification module}, which compares sub-symbolic representations of logic atoms, and mutually recursive $\module{or}$ and $\module{and}$ modules, which jointly enumerate all possible proof paths, before the final aggregation selects the highest scoring one.
The whole process is outlined in \cref{alg:bc}.
\begin{example}[Neural Theorem Provers] \label{ex:2}
Consider a variant of \cref{ex:1}, where each predicate and constant lives in a continuous embedding space, \ie{} $\params_{\rel{p}:}, \params_{\rel{q}:}, \params_{\const{rick}:}, \params_{\const{beth}:}, \params_{\const{morty}:} \in \Real^{k}$.
\citet{DBLP:conf/nips/Rocktaschel017} propose replacing comparisons between symbols with a differentiable similarity measure $\krnl : \Real^{k} \times \Real^{k} \rightarrow \left[ 0, 1 \right]$, such as a Gaussian kernel, between their embeddings.
Their model enumerates all possible proofs for a goal $G$, and generates a \emph{proof score} for each of them, given by the \emph{minimum} of all embedding similarities.
For instance, if $G = \rel{grandPa}(\const{rick}, \const{morty}) = \xs{\rel{grandPa}, \const{rick}, \const{morty}}$, one candidate proof consists in using the facts $F~=~\rel{p}(\const{rick}, \const{beth})$ and $F'~=~\rel{p}(\const{beth}, \const{morty})$ and the rule $\rel{g}(\var{X}, \var{Y}) \lif \rel{p}(\var{X}, \var{Z}), \rel{p}(\var{Z}, \var{Y})$ from the \gls{KB}, yielding the score $\krnl(\params_{\rel{grandPa}}, \params_{\rel{g}})$.
It is important to mention that \glspl{NTP} allow unifying symbols like $\rel{grandPa}$ and $\rel{g}$ (which, in this case, share the same semantics), even though they are lexically different.
The score for $G$ is given by the \emph{maximum} of all proof scores. $\triangle$
\end{example}
\begin{algorithm}[t]
    \caption{In Conditional Theorem Provers, the set of rules is conditioned on the goal $G$.} \label{alg:or}
    \begin{algorithmic}[1]
        \Function{$\module{or}$}{$G$, $d$, $\state$} \label{ln:new_or}
            \For{$\lst{H} \lif \lss{B} \in \select_{\theta}(G)$} \label{ln:some}
                \For{$\state \in \module{and}\left(\lss{B}, d, \module{unify}(\lst{H}, \lst{G}, \state)\right)$}
                    \State {\bf yield} $\state$
                \EndFor
            \EndFor
        \EndFunction
    \end{algorithmic}
\end{algorithm}
\section{Conditional Proving Strategies} \label{sec:conditional}
The \glspl{NTP} proposed by \citet{DBLP:conf/nips/Rocktaschel017} use a fixed set of rules, either specified by the user or learned from data via provided rule templates.
In this model, given a goal, there is no \emph{hard} decision mechanism for deciding which rules can be used for reformulating a given goal into subgoals: all rules in the \gls{KB} need to be considered when proving each goal.
For this reason, \glspl{NTP} were shown not to scale to large datasets~\citep{DBLP:conf/nips/Rocktaschel017}.
\paragraph{Differentiable Goal Reformulation}
In order to explicitly \emph{learn} which rule to consider at each step, we propose the following solution.
Rather than relying on a fixed, potentially very large set of rules, we propose to dynamically \emph{generate} a minimal set of rules via a neural network architecture conditioned on the goal to prove.
\begin{example}[Conditional Theorem Proving]
Assume that the goal to prove is $G = \rel{g}(\const{rick}, \const{morty})$, we want the model to be able to only consider the best rules for proving $G$, such as $\rel{g}(\var{X}, \var{Y}) \lif \rel{p}(\var{X}, \var{Z}), \rel{p}(\var{Z}, \var{Y})$, rather than all rules in the \gls{KB}.
Remember that, in \glspl{NTP}, relations and constants in the \gls{KB} are represented by their embedding vectors, and the aforementioned rule selection can be implemented via a mapping from $\params_{\rel{g}:}$ to $\xs{\params_{\rel{p}:}, \params_{\rel{p}:}}$.
$\triangle$
\end{example}
Consider the $\module{or}$ module in \cref{alg:bc}.
Selecting which rule to use during the proving process for a given goal $G$ can be implemented by rewriting the $\module{or}$ module as in \cref{alg:or}, where the set of clauses is produced by a module that, given the goal, generates a set of rules for proving $G$.
The core difference between \glspl{NTP} and \glspl{CTP} is that, when proving a goal $G$, rather than iterating through a possibly very large set of clauses in the \gls{KB} $\kb$ (see \cref{ln:all} in the $\module{or}$ module definition in \cref{alg:bc}), the conditional $\module{or}$ module in \cref{alg:or} only iterates through a small set of generated clauses, whose generation is conditioned on $G$ (see \cref{ln:some} in \cref{alg:or}).
Given a goal $G$, the $\select$ module with parameters $\theta$ in \cref{alg:or} produces a set of clauses, each specifying which sub-goals to prove in order to produce a proof for $G$.
Note that the $\select$ module can be implemented by an end-to-end differentiable parametrised function $f(\cdot)$ that, given a goal $G$, produces a finite sequence of corresponding subgoals:
\begin{equation} \label{eq:signature}
\begin{aligned}
\select_{\theta}(G) : \mathcal{A} \rightarrow & \left[ \mathcal{A} \lif \mathcal{A}^{*} \right],
\end{aligned}
\end{equation}
\noindent where $\mathcal{V}$ is a set of variables, and $\mathcal{A} \triangleq \Real^{k} \times (\Real^{k} \cup \mathcal{V}) \times (\Real^{k} \cup \mathcal{V})$ denotes the embedding representation of a goal, such as $\rel{g}(\const{rick}, \const{morty})$. 
For instance, the $\select$ module in \cref{eq:signature} can be implemented by a neural network that, given a goal such as $G = \xs{\params_{\rel{g}:}, \params_{\const{rick}:}, \params_{\const{morty}:}}$, generates $\lst{H} \lif \lst{B}$ with $\lst{H} = \xs{\params_{g:}, \var{X}, \var{Y}}$ and $\lst{B} = \xs{\xs{\params_{p:}, \var{X}, \var{Z}}, \xs{\params_{p:}, \var{Z},\var{Y}}}$, corresponding to the symbolic rule $\rel{g}(\var{X}, \var{Y}) \lif \rel{p}(\var{X}, \var{Z}), \rel{p}(\var{Z}, \var{Y})$.
If the positions of the variables in the rule are fixed, the whole module is end-to-end differentiable with respect to its parameters $\theta$.

\paragraph{Neural Goal Reformulation}
Here, we define $\select$ as a linear function of the goal predicate:
\begin{equation} \label{eq:linear}
\begin{aligned}
\select_{\theta}(G) \triangleq & \left[ F_{\lst{H}}(G) \lif F_{\lst{B}_{1}}(G), F_{\lst{B}_{2}}(G) \right],
\end{aligned}
\end{equation}
\noindent where the head and body of the resulting rule are given by $F_{\lst{H}}(G) = \xs{ f_{\lst{H}}(\params_{G_{1}}), \var{X}, \var{Y} }$, $F_{\lst{B}_{1}}(G) = \xs{ f_{\lst{B}_{1}}(\params_{G_{1}}), \var{X}, \var{Z} }$, and $F_{\lst{B}_{2}}(G) = \xs{ f_{\lst{B}_{2}}(\params_{G_{1}}), \var{Z}, \var{Y} }$.
Every $f_{i} : \Real^{k} \rightarrow \Real^{k}$ is a differentiable function, such as the linear projection $f_{i}(\mathbf{x}) = \mathbf{W}_{i} \mathbf{x} + \mathbf{b}$, with $\mathbf{W}_{i} \in \Real^{k \times k}$ and $\mathbf{b} \in \Real^{k}$.
%
%
Thus, instead of iterating through a possibly very large set of rules in the \gls{KB} $\kb$, we can generate a significantly smaller set of rules, whose generation is conditioned on the goal $G$ and can be trained end-to-end on downstream reasoning tasks.
\paragraph{Attentive Goal Reformulation}
We can incorporate a useful prior in the \module{select} module architecture---namely that predicate symbols in the rule already exist in the \gls{KB}, among the available relations $\rels$.
A method for incorporating this prior consists in using the given goal $G$ for generating a distribution over the set of relations $\rels$:
\begin{equation} \label{eq:attentive}
\begin{aligned}
f_{i}(\mathbf{x}) & = \attn \mathbf{E}_{\rels} \\
\attn & = \softmax\left( \mathbf{W}_{i} \mathbf{x}\right) \in \Delta^{|\rels| - 1},
\end{aligned}
\end{equation}
\noindent where $\mathbf{E}_{\rels} \in \Real^{|\rels| \times k}$ denotes the predicate embedding matrix, $\mathbf{W}_{i} \in \Real^{k \times |\rels|}$, and $\attn$ is an attention distribution $\attn \in \Delta^{|\rels| - 1}$ over the predicates in $\rels$, where $\Delta^{n}$ denotes the standard $n$-simplex.\footnote{The standard $n$-simplex $\Delta^{n}$ is defined as $\Delta^{n} = \{ (\attn_{0}, \ldots, \attn_{n}) \in \Real^{n + 1} \mid \sum_{i = 0}^{n} \attn_{i} = 1 \land \forall i: \attn_{i} \geq 0 \}$}
This is especially helpful if the embedding size is larger than the number of relationships, \ie{} $k \gg |\rels|$.
\paragraph{Memory-Based Goal Reformulation}
A problem with using black-box neural networks for reformulating goals into sub-goals is that it can be difficult to inspect the rules by analysing the model parameters, as in \citet{DBLP:conf/nips/Rocktaschel017}.
For this reason, we propose an alternative goal reformulation module, where rules are stored in a differentiable memory.
More precisely, $n$ rules are stored as memory matrices $\left[\mathbf{M}_1,\ldots,\mathbf{M}_m\right]$, where each $\mathbf{M}_i \in \Real^{n \times k}$ denotes the $k$-dimensional embeddings of the $i$-th predicates in the $n$ rules.
Then, the goal $G$ is used to compute an attention distribution over rules $\attn \in \Delta^{n - 1}$, where each $\attn_{i}$ denotes the attention weight on the $i$-th rule.
The attention distribution can be formalised as follows:
\begin{equation} \label{eq:memory}
\begin{aligned}
f_{i}(\mathbf{x}) & = \attn \mathbf{M}_{i} \\
\attn & = \softmax\left( \mathbf{W} \mathbf{x} \right) \in \Delta^{n - 1},
\end{aligned}
\end{equation}
\noindent where each $f_{i} : \Real^{k} \rightarrow \Real^{k}$ is a differentiable function that, given the goal, produces an attention distribution $\attn \in \Delta^{n - 1}$ over the rules and for indexing a memory $\mathbf{M}_{i}$, analogous to a key-value memory network~\citep{DBLP:conf/emnlp/MillerFDKBW16}.
\begin{figure}[t!]
%
%
\centering
\includegraphics[width=0.95\columnwidth]{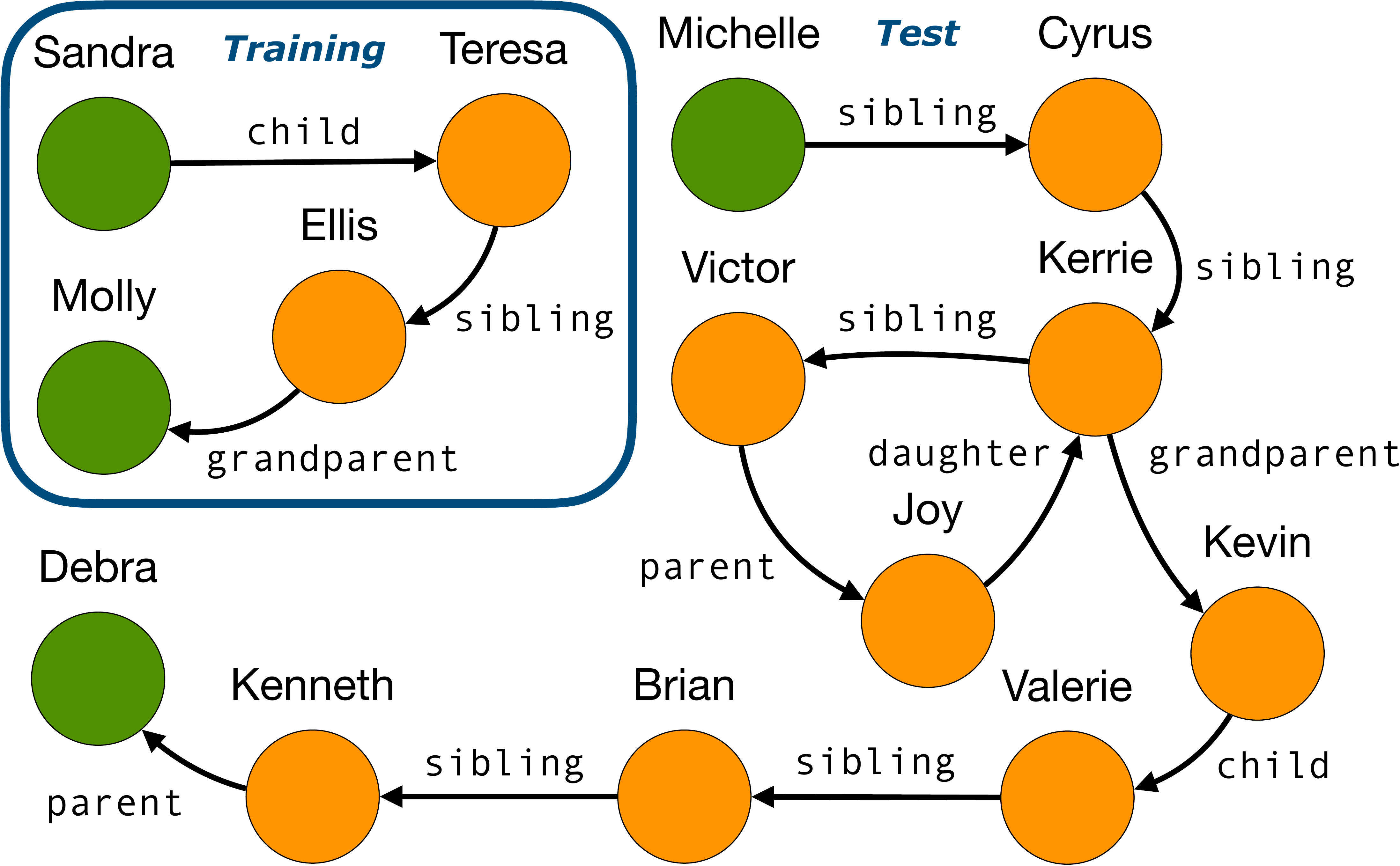}
\caption{Example of a train and test instance in \glsentryshort{CLUTRR} -- the training instance (upper left) is composed by a graph with three edges, while the test instance is composed by ten edges; the task consists in identifying the relationships between the green nodes.} \label{fig:clutrr}
%
%
%
%
\end{figure}
%

%
%

%
\section{Related Work} \label{sec:related}
%
%
\paragraph{Memory-Augmented Networks}
Memory-augmented neural architectures aim at improving the generalisation and reasoning abilities in neural networks by disentangling representations from computations.
By enriching neural networks with a differentiable \emph{external memory}, these models were able to multi-hop reason over texts~\citep{DBLP:conf/nips/SukhbaatarSWF15}, induce algorithmic behaviours~\citep{DBLP:journals/corr/GravesWD14,joulin2015inferring,grefenstette2015learning,DBLP:journals/corr/KaiserS15}, and rapidly assimilate new data~\citep{santoro2016meta}.
%

\input{tables/rules}

%
%
\paragraph{Neuro-Symbolic Models}
\emph{Differentiable interpreters} enable translating declarative or procedural knowledge into neural networks exhibiting strong inductive biases of that knowledge~\citep{DBLP:conf/icml/BosnjakRNR17,DBLP:conf/nips/Rocktaschel017,evans2018learning}.
\citet{DBLP:conf/icml/BosnjakRNR17} propose $\partial 4$, a differentiable abstract machine for the Forth programming language.
\citet{DBLP:conf/nips/Rocktaschel017} propose a differentiable implementation for the backward chaining algorithm, effectively implementing a differentiable Datalog interpreter.
\citet{evans2018learning} propose a differentiable forward-chaining reasoning process, while \citet{DBLP:conf/ijcai/DonadelloSG17} propose a continuous generalisation of the semantics of first-order logic.
\citet{DBLP:conf/nips/YangYC17} and \citet{DBLP:conf/nips/SadeghianADW19} propose an approach for learning function-free Datalog clauses from \glspl{KB} by means of a differentiable graph traversal operator, while \citet{DBLP:conf/iclr/DasDZVDKSM18} propose learning policies for navigating a \gls{KB} via reinforcement learning.
A major problem with these approaches is their computational complexity, which renders them unusable for larger-scale learning problems. 
In order to address this issue, \citet{DBLP:conf/aaai/MinerviniBR0G20} propose \glspl{GNTP}, an extension to \glspl{NTP} where, for each goal, only the top-$k$ facts and rules are considered during the differentiable reasoning process.
%

%
%
%

%
\paragraph{Neural Module Networks}
\citet{DBLP:conf/cvpr/AndreasRDK16} introduce \glspl{NMN}, an end-to-end differentiable composition of jointly trained neural modules.
Analogously, \glspl{NTP} can be seen as a recursive differentiable composition of \module{or} and \module{and} modules, jointly trained on downstream reasoning tasks.
\glspl{NMN} allow defining and training end-to-end differentiable composable models, and interpret and execute their compositions as simple programs.
This is especially useful when dealing with reasoning tasks from visual and natural language inputs, such as reasoning over text with arithmetic modules~\citep{DBLP:journals/corr/abs-1912-04971} or visual question answering~\citep{DBLP:conf/cvpr/AndreasRDK16}.
Interestingly, in this work the structure of the composition is statically drawn from the data, while \citet{DBLP:conf/emnlp/JiangB19} propose a way of learning the model composition via a coordination model~\citep{DBLP:conf/emnlp/JiangB19}.
\paragraph{Incorporating Knowledge via Regularisation}
Another branch of works uses symbolic background knowledge to learn better representations for entities and relationships in a \gls{KB}.
An early work in this space is \citet{DBLP:conf/naacl/RocktaschelSR15}, which regularise a relation extraction model by penalising inconsistency with respect to a set of logical constraints and sampled entities.
\citet{DBLP:conf/pkdd/MinerviniCMNV17} regularise relation representations to incorporate equivalence and inversion axioms for a set of neural link prediction models, while \citet{DBLP:conf/emnlp/DemeesterRR16} focus on simple implication axioms.
\citet{DBLP:conf/uai/MinerviniDRR17} propose adversarially regularising neural models by identifying, during training, inputs that violate a given set of constraints, and regularising the model to decrease the degree of such violations.
\citet{DBLP:conf/icml/XuZFLB18} propose a similar idea, by using a \emph{semantic loss} measuring to which extent a model matches a set of given constraints.
%

\input{figures/089907f8}

\input{figures/db9b8f04}

%
\section{Experiments} \label{sec:exp}
We evaluate \glspl{CTP} on two datasets: systematic generalisation on the \gls{CLUTRR} dataset, and link prediction in \glspl{KG}.
Datasets are introduced in \cref{ssec:datasets}, while baselines are described in \cref{ssec:baselines}. 
\subsection{Datasets and Tasks} \label{ssec:datasets}
\paragraph{Systematic Generalisation}
\gls{CLUTRR}---Compositional Language Understanding and Text-based Relational Reasoning~\citep{DBLP:conf/emnlp/SinhaSDPH19}---contains a large set of graphs modelling hypothetical family relationships.
Given a set of family relations, encoded as a graph with a variable number of nodes and edges, the goal is to infer the relationship between two family members, whose relationship is not explicitly mentioned.
To solve this task, a learning agent should be able to induce the logical rules governing the kinship relationships, such as the \emph{parent} of a \emph{parent} is a \emph{grandparent}, and use these rules to infer the relationship between a given pair of entities.
\gls{CLUTRR} allows testing a learning agent's ability for \emph{systematic generalisation}, by testing on graphs containing combinations of logical rules that were not seen during training.
%
%
Each edge in the graph is labelled with one out of nine family relation type from $\rels = \{$ child, grand, in-law, inv-child, inv-grand, inv-in-law, inv-un, sibling, un $\}$, and the task consists in inferring the relationship between two of the nodes in the graph.
During training, a model is trained to infer such relationship by traversing a limited number of edges (such as two, three, and four edges), and during evaluation the model has to traverse up to ten edges.
\cref{fig:clutrr} shows an example of a training instance and a test instance in \gls{CLUTRR}: the training instances consists in a graph modelling a set of family relations of only three edges, while the test instance is composed by a graph with ten edges.
In both cases, the task consists in inferring the relationships between two of the nodes in the graph.
\paragraph{Link Prediction}
Furthermore, we evaluate \glspl{CTP} on neural link prediction tasks,  following the same evaluation protocols as \citet{DBLP:conf/nips/Rocktaschel017} on the Countries~\citep{bouchard2015approximate}, Nations, UMLS, and Kinship~\citep{kemp2006learning} datasets.
The Countries dataset contains countries, regions, and sub-regions as entities, and it is carefully designed to test the logical reasoning and learning capabilities of neural link prediction models: queries have the form $\rel{locatedIn}(\const{c}, {}\cdot{})$, where the answers are regions. This dataset comes with three tasks ($S1$, $S2$, and $S3$) each requiring reasoning skills of increasing complexity; we refer to \citet{DBLP:conf/nips/Rocktaschel017} for more details about this dataset.
The Unified Medical Language System (UMLS) dataset is from bio-medicine: entities are biomedical concepts, and relations include treatments and diagnoses.
The Kinship dataset contains kinship relationships among members of the Alyawarra tribe from Central Australia.
\subsection{Models and Baselines} \label{ssec:baselines}
\paragraph{Models}
We consider the following three \gls{CTP} model variants:
\begin{inparaenum}[\itshape i\upshape)]
\item CTP$_{L}$, where the mappings $f_{i}$ from goal predicates to rule predicates are implemented via a linear projection,
\item CTP$_{A}$, where it is implemented via \emph{attentive goal reformulation}, as described in \cref{sec:conditional}, and
\item CTP$_{M}$, where it is implemented via \emph{memory-based goal reformulation}, also described in \cref{sec:conditional}.
\end{inparaenum}
\paragraph{Baselines} We consider two classes of baselines: \emph{graph-based} and \emph{sequence-based}.
Graph-based baselines consist in neural architectures specifically designed for graph data and. We consider \glspl{GNTP}, a neuro-symbolic reasoning model, and two \gls{GNN} architectures, namely \glspl{GAT}~\citep{DBLP:conf/iclr/VelickovicCCRLB18} and \glspl{GCN}~\citep{DBLP:conf/iclr/KipfW17}.
Sequence-based baselines are neural architectures originally proposed for encoding sequences: by linearising the relational graphs into sequences of subject-predicate-object triples, we can use such models for encoding graphs.
We consider several sequence encoding models, namely
\glspl{RNN},
\glspl{LSTM}~\citep{DBLP:journals/neco/HochreiterS97},
\glspl{GRU}~\citep{DBLP:conf/emnlp/ChoMGBBSB14},
\glspl{CNN}~\citep{DBLP:conf/emnlp/Kim14},
\glspl{CNNH}~\citep{DBLP:conf/aaai/KimJSR16},
\glspl{MHA}~\citep{DBLP:conf/nips/VaswaniSPUJGKP17}.
\paragraph{Encoding}
For both graph-based and sequence-based baselines, we considered two approaches:
\begin{inparaenum}[\itshape i\upshape)]
\item encoding the the \gls{KB} and goal independently, as in \citet{DBLP:conf/emnlp/SinhaSDPH19}, and \label{enc:ind}\item conditioning the \gls{KB} encoding on the goal. \label{enc:dep}
\end{inparaenum}
Let $\encoder_{\params_{e}}$ denote an encoder that, given a set of ground facts (such as a \gls{KB} or a goal), produces a continuous $k$-dimensional representation, and $\hat{y}$ denote a conditional distribution over the candidate relationship types.
The encoder in item (\ref{enc:ind}), where the goal $G$ and the \gls{KB} $\kb$ are encoded independently, can be summarised as $\hat{y} = \softmax(\mathbf{W} \left[ \encoder_{\params_{e}}(\kb) ; \encoder_{\params_{e}}(G) \right])$.
The encoder in item (\ref{enc:dep}), where $G$ and $\kb$ are encoded jointly, can be summarised as $\hat{y} = \softmax(\mathbf{W} \encoder_{\params_{e}}(\left[ G ; \kb \right]))$.

For model selection, we generate a \gls{CLUTRR}-like dataset using the code published by \citet{DBLP:conf/emnlp/SinhaSDPH19} composed of training set graphs with $\{2, 3\}$ edges, and two validation sets, one with graphs with three edges, and another with graphs with nine edges.
We then select two sets of hyperparameters for each of the baselines: one that maximises the validation accuracy on graphs with three edges, and another that maximises the test accuracy on graphs with nine edges.
All details on the hyperparameter selection process can be found in \cref{app:hyper}.
To assess the statistical significance of our results, we ran each of the experiments $10$ times, each time with a different seed, and compared the resulting accuracy values using an \emph{unequal variances t-test}, or Welch's $t$-test.\footnote{We assume accuracy values to be Gaussian-distributed, as they 
approach a normal distribution for large numbers of re-runs, 
due to the Central Limit Theorem.}
\subsection{Results} \label{ssec:results}
\paragraph{CLUTRR}
We evaluated three \gls{CTP} variants and all considered baselines on two datasets published by \citet{DBLP:conf/emnlp/SinhaSDPH19} under the identifiers \texttt{089907f8} and \texttt{db9b8f04}---we refer to these datasets as $\text{CLUTRR}_\text{G}(k=2,3)$ and $\text{CLUTRR}_\text{G}(k=2,3,4)$, where $k$ denotes the number of edges in the training graphs.
Results for $\text{CLUTRR}_\text{G}(k=2,3)$ are summarised in \cref{fig:089907f8}, while results for $\text{CLUTRR}_\text{G}(k=2,3,4)$ are summarised in \cref{fig:db9b8f04}.
In \cref{fig:089907f8}, we observe that, after training on graphs with two and three edges, baseline models tend to be able to generalise correctly to slightly longer stories (such as graphs with four and five edges), but that predictive accuracy quickly decreases with increasing graph sizes and this phenomenon happens when tuning hyper-parameters either on graphs with three edges or on graph with nine edges.
In our experiments, \glspl{LSTM} had a strikingly different behaviour in comparison with other baselines: for graphs with nine edges, the accuracy decrease caused by using the \glspl{LSTM} baseline is only significant with $p \leq 10^{-2}$ (for all other baselines this change is significant with $p \leq 10^{-4}$), with a drop in significance for smaller graphs.
The phenomenon that \glspl{LSTM} yield surprisingly accurate results on the \gls{CLUTRR} dataset can be seen across every experiment in our empirical evaluation, while other recurrent models such as \glspl{RNN} and \glspl{GRU} do not show this.
\begin{figure}
    \centering
    \includegraphics[width=\columnwidth]{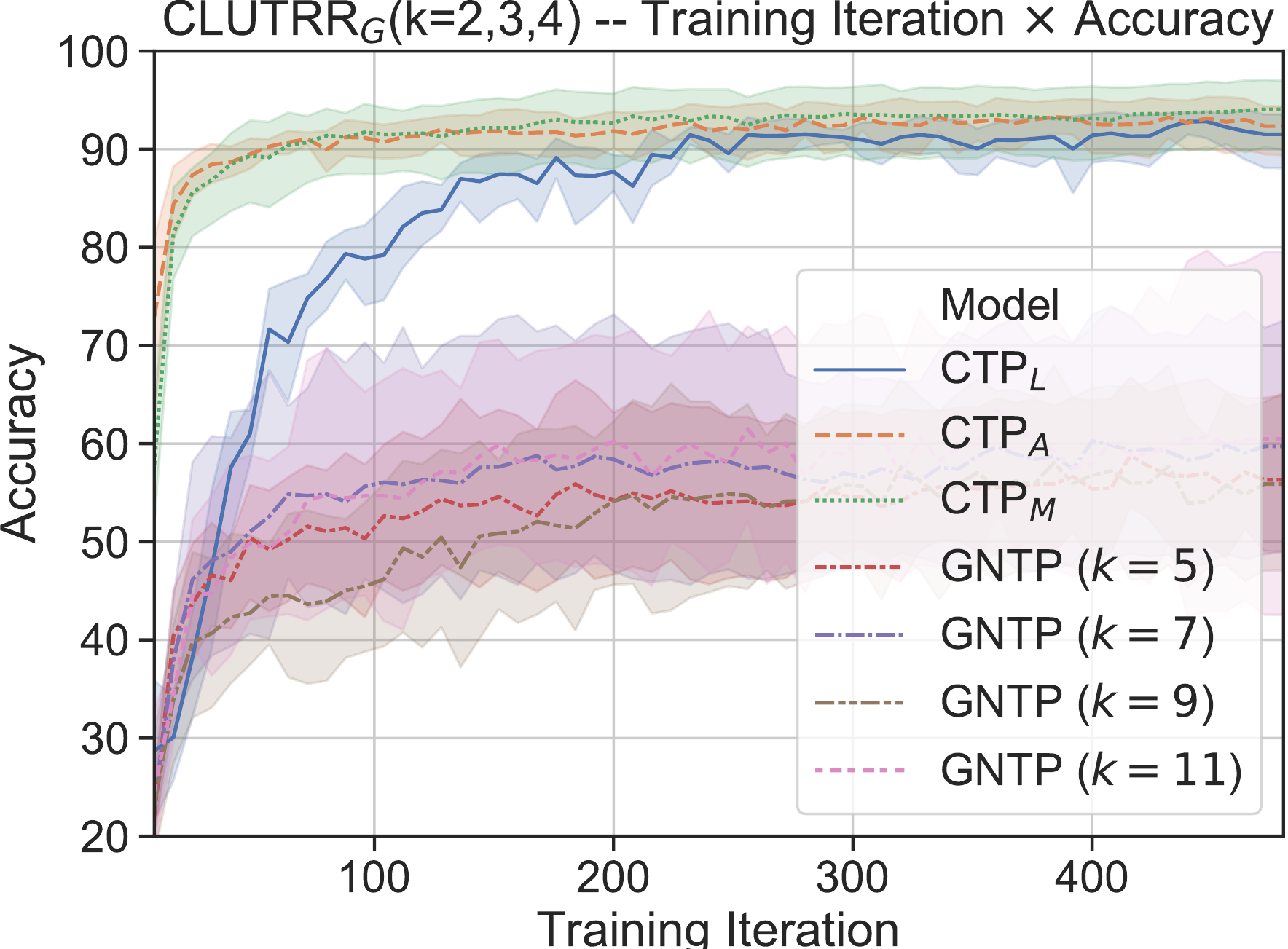}
    \caption{Training dynamics of \glspl{CTP} and \glspl{GNTP}.}
    \label{fig:dynamics}
\end{figure}
\input{tables/small_rules}
\paragraph{Model Analysis}
A great feature of \glspl{CTP} is that we can analyse the goal reformulation process to understand the reasoning process underlying a given prediction, and extract explainable rules.
In \cref{tab:clutrr-rules}, we show a sample of the rules and common-sense reasoning patterns learned by \glspl{CTP} on the \gls{CLUTRR} dataset.
We can see that, for example, \glspl{CTP} successfully identify that \eg{} \emph{the child of a child is a grandchild}, \emph{the child of one's significant other is also one's child}, and \emph{the parent of a significant other is an in-law}.
\paragraph{Training Dynamics}
We analyse the training dynamics of \gls{CTP}$_{L}$, \gls{CTP}$_{A}$, \gls{CTP}$_{M}$, and \glspl{GNTP}~\citep{DBLP:conf/aaai/MinerviniBR0G20} on CLUTRR$_G$(k=2,3,4).
The \gls{CTP} variants consisted of 5 goal reformulators, each implemented by an independent $\module{select}$ module, while \gls{GNTP} has a \gls{KB} of 32 rules and $k \in \{ 5, 7, 9, 11 \}$. For all models, the embedding size for entities and relations was set to 50.
\Cref{fig:dynamics} demonstrates how the training accuracy of such models evolves during training.
We can see that, while the three \gls{CTP} variants get to a nearly-perfect training set accuracy in less than 300 iterations, \glspl{GNTP} is unable to match this result, even after careful hyperparameter tuning.
A possible explanation is that, in order to scale to large rule sets, \glspl{GNTP} only considers the top-$k$ rules, based on the similarity between the goal and the head of the rules.
This is equivalent to a hard attention mask, which is known to be problematic to train via gradient-based optimisation~\citep{DBLP:conf/emnlp/LuongPM15}.
\paragraph{Link Prediction}
In \cref{tab:benchmark}, we show link prediction results in comparison with three other neuro-symbolic reasoning methods, namely \glspl{GNTP}~\citep{DBLP:conf/aaai/MinerviniBR0G20}, NeuralLP~\citep{DBLP:conf/nips/YangYC17} and MINERVA~\citep{DBLP:conf/iclr/DasDZVDKSM18}.
\glspl{GNTP} are an extension of \glspl{NTP} where rules are heuristically selected by search for the rules where the head predicate is closest to the sub-goal predicate in embedding space.
Our experiments show that \glspl{CTP} produce significantly more accurate or very competitive link prediction results, while controlling the complexity of the reasoning process via the goal-conditioned rule selection.
For instance, in the Nations dataset, only two rules were generated by \glspl{CTP} for each goal, while in \citet{DBLP:conf/nips/Rocktaschel017} \glspl{NTP} were required to iterate over sixty rules.
Furthermore, in this case, \glspl{CTP} were able to produce explanations for each of their predictions.
For instance, in the Nations dataset, \glspl{CTP} successfully extracted logical patterns such as $\rel{commonbloc1}(\var{X}, \var{Y}) \lif \rel{relngo}(\var{Y}, \var{X})$, $\rel{timesincewar}(\var{X}, \var{Y}) \lif \rel{independence}(\var{X}, \var{Y})$, $\rel{unweightedunvote}(\var{X}, \var{Y}) \lif \rel{relngo}(\var{X}, \var{Y})$, and $\rel{ngo}(\var{X}, \var{Y}) \lif \rel{independence}(\var{Y}, \var{X})$.
\section{Conclusions} \label{sec:end}
We introduced \glspl{CTP}, an extension to \glspl{NTP} for learning the optimal rule selection strategy via gradient-based optimisation. 
For each sub-goal, a \module{select} module produces a smaller set of rules, which is then used during the proving mechanism.
Furthermore, we proposed three variants of the rule selection mechanism, where the sub-goal reformulations are obtained by linear projections of the sub-goal predicate, attention distributions over predicate embeddings, and a key-value memory lookup over a set of rules.
We showed that \glspl{CTP} are scalable and yield state-of-the-art results on the \gls{CLUTRR} dataset, which explicitly tests the systematic generalisation of neural models, in comparison with a wide set of neural baselines.
Finally, we demonstrated that \glspl{CTP} yield competitive results in standard link prediction benchmark in comparison with other neuro-symbolic approaches.
\paragraph{Future Work} An open problem is how \glspl{CTP} can be able to process \gls{CLUTRR} instances where family relationships are not directly provided as a labelled graph, but rather as free-form text.
A possible solution, proposed by \citet{DBLP:conf/aaai/MinerviniBR0G20}, consists in having an end-to-end differentiable encoder for producing the fact embeddings conditioned on the text, and we are currently analysing several options in this space.

\paragraph{Acknowledgements}
This work was supported by the EU Horizon 2020 Research and Innovation Programme under the grant 875160. We thank Yihong Chen, Joe Stacey, and all the amazing folks in the UCL NLP group for the enlightening discussions and support. Finally, we thank NVIDIA for GPU donations.

\bibliographystyle{icml2020}
\bibliography{bibliography}

\clearpage

\appendix

\section{Hyperparameter Selection} \label{app:hyper}
For model selection in baseline models, we generated a \gls{CLUTRR}-like dataset using the code published by \citet{DBLP:conf/emnlp/SinhaSDPH19} composed by a training set of graphs with $\{2, 3\}$ edges, and two test sets, one with graphs with three edges, and another with graphs with nine edges.
We then selected two sets of hyperparameters for each of the baselines: one that maximises the validation accuracy on graphs with three edges, and another that maximises the test accuracy on graphs with nine edges.
For each of the baselines, we considered a wide range of hyperparameters: the dimensionalities of node and edge embeddings were varied in $\{ 10, 50, 100, 200, 500 \}$, the number of attention heads in attention-based architectures in $\{ 1, 2, \ldots, 10 \}$, the number of filters in convolutional architectures in $\{ 1, 2, \ldots, 10 \}$, and the number of hidden units in recurrent architectures in $\{ 32, 64, 128, 256, 512 \}$.
To assess the statistical significance of our results, we ran each of the experiments $10$ times, each time with a different seed, and compared the resulting accuracy values using an unequal variances t-test, or Welch's $t$-test.
This is motivated by the observation that accuracy values to be Gaussian-distributed, as they 
approach a normal distribution for large numbers of re-runs, 
due to the Central Limit Theorem.

\subsection{Optimal Hyperparameters}
Note that all recurrent architectures are bidirectional.
\begin{description}
\item[Graph Attention Networks:] the hyperparameters that maximise the accuracy on validation graphs with 3 edges are $h = 10$ for the number of attention heads, $k = 50$ for the dimensionality of node embeddings, $k_{e} = 200$ for the dimensionality of edge embeddings. For validation graphs with 9 edges, $k = 50$, $k_{e} = 500$, and $h = 10$.
\item[Graph Convolutional Networks:] the hyperparameters that maximise the accuracy on validation graphs with 3 edges are $k = 50$ for the dimensionality of node embeddings, $k_{e} = 500$ for the dimensionality of edge embeddings. For validation graphs with 9 edges, $k = 50$, and $k_{e} = 50$.
\item[Convolutional Neural Networks:] the hyperparameters that maximise the accuracy on validation graphs with 3 edges are $k = 50$ for the dimensionality of node and edge embeddings, $f = 8$ convolutional filters, and conditional encoding. For validation graphs with 9 edges, $k = 200$, $f = 4$, and conditional encoding.
\item[Recurrent Neural Networks:] the hyperparameters that maximise the accuracy on validation graphs with 3 edges are $k = 50$ for the dimensionality of node and edge embeddings, $h = 64$ for the size of the hidden state representations, and conditional encoding. For validation graphs with 9 edges, $k = 500$, $h = 512$, and conditional encoding.
\item[Long Short-Memory Networks:] the hyperparameters that maximise the accuracy on validation graphs with 3 edges are $k = 50$ for the dimensionality of node and edge embeddings, $h = 64$ for the size of the hidden state representations, and conditional encoding. For validations graphs with 9 edges, $k = 100$, $h = 512$, and independent encoding.
\item[Gated Recurrent Units:] the hyperparameters that maximise the accuracy on validation graphs with 3 edges are $k = 50$ for the dimensionality of node and edge embeddings, $h = 64$ for the size of the hidden state representations, and conditional encoding. For validation graphs with 9 edges, $k = 200$, $h = 512$, and conditional encoding.
\item[CNN with Highway Encoder:] the hyperparameters that maximise the accuracy on validation graphs with 3 edges are $k = 200$ for the dimensionality of node and edge embeddings, $h = 2$ highway layers, and conditional encoding. For validation graphs with 9 edges, $k = 200$, $h = 1$, and conditional encoding.
\item[Multi-Head Attention:] the hyperparameters that maximise the accuracy on validation graphs with 3 edges are $k = 500$ for the dimensionality of node and edge embeddings, $h = 10$ for the number of attention heads, $h_{k}$ for the size of the hidden state representation of the top \gls{LSTM} layer, and conditional encoding. For validation graphs with 9 edges, $k = 500$, $h = 10$, $h_{k} = 128$, and conditional encoding.
\end{description}

\end{document}

%% file: math_commands.tex
\usepackage{amsmath,amsfonts,bm}

\def\eqref#1{equation~\ref{#1}}

\def\1{\bm{1}}

\DeclareMathAlphabet{\mathsfit}{\encodingdefault}{\sfdefault}{m}{sl}
\SetMathAlphabet{\mathsfit}{bold}{\encodingdefault}{\sfdefault}{bx}{n}

\newcommand{\softmax}{\mathrm{softmax}}



%% file: packages.tex
\usepackage{ifxetex}

\ifxetex
 \usepackage{fontspec}
\else
 \usepackage[utf8]{inputenc}
\fi

\usepackage{enumitem}
\usepackage[usenames,dvipsnames]{xcolor}

\usepackage{multicol}

\definecolor{nice-red}{HTML}{E41A1C}
\colorlet{dark-red}{nice-red!80!black}
\definecolor{nice-orange}{HTML}{FF7F00}
\colorlet{dark-orange}{orange!85!black}
\definecolor{nice-yellow}{HTML}{FFC020}
\definecolor{nice-green}{HTML}{4DAF4A}
\definecolor{nice-blue}{HTML}{377EB8}
\definecolor{nice-purple}{HTML}{984EA3}

\usepackage{url}            

\usepackage[xindy]{glossaries}

\usepackage[]{todonotes}

\usepackage{bm}
\usepackage[capitalise]{cleveref}

\crefformat{equation}{Eq.~#2#1#3}
\Crefformat{equation}{Equation~#2#1#3}

\crefrangeformat{equation}{Eqs.~#3#1#4 to~#5#2#6}
\Crefrangeformat{equation}{Equations~#3#1#4 to~#5#2#6}
\crefmultiformat{equation}{Eqs.~#2#1#3}%
{ and~#2#1#3}{, #2#1#3}{ and~#2#1#3}
\Crefmultiformat{equation}{Equations~#2#1#3}%
{ and~#2#1#3}{, #2#1#3}{ and~#2#1#3}

\usepackage{tikz}
\usetikzlibrary{fit,shapes.misc}

\usetikzlibrary{calc,trees,positioning,arrows,chains,shapes.geometric,%
  decorations.pathreplacing,decorations.pathmorphing,shapes,%
  matrix,shapes.symbols,fit,decorations,arrows.meta}
 
\usepackage{wrapfig}

\usepackage{comment}

\usepackage{amsmath}
\usepackage{amsthm}

\theoremstyle{definition}
\newtheorem{example}{Example}[section]

\usepackage{multirow}

\usepackage{amssymb}
\usepackage{soul}

\usepackage{quoting}
\quotingsetup{vskip=5pt,indentfirst=false,leftmargin=5pt,rightmargin=5pt}

\usepackage{paralist}

\usepackage{tabularx}

\usepackage{lscape}

\newcolumntype{b}{X}
\newcolumntype{s}{>{\hsize=.2\hsize}X}
\newcolumntype{m}{>{\hsize=.4\hsize}X}
\newcolumntype{z}{>{\hsize=.8\hsize}X}

\usepackage{rotating}

\usepackage{array}
\newcolumntype{L}[1]{>{\raggedright\let\newline\\\arraybackslash\hspace{0pt}}m{#1}}
\newcolumntype{C}[1]{>{\centering\let\newline\\\arraybackslash\hspace{0pt}}m{#1}}
\newcolumntype{R}[1]{>{\raggedleft\let\newline\\\arraybackslash\hspace{0pt}}m{#1}}



%% file: acronyms.tex
\newacronym{NTP}{NTP}{Neural Theorem Prover}
\newacronym{NLU}{NLU}{Natural Language Understanding}
\newacronym{QA}{QA}{Question Answering}
\newacronym{NLI}{NLI}{Natural Language Inference}

\newacronym{KB}{KB}{Knowledge Base}
\newacronym{KG}{KG}{Knowledge Graph}

\newacronym{NLP}{NLP}{Natural Language Processing}

\newacronym{GNN}{GNN}{Graph Neural Network}

\newacronym{GAT}{GAT}{Graph Attention Network}
\newacronym{GCN}{GCN}{Graph Convolutional Network}

\newacronym{BOE}{BOE}{Bag of Embeddings}
\newacronym{CNN}{CNN}{Convolutional Neural Network}
\newacronym{CNNH}{CNNH}{CNN with Highway Encoder}

\newacronym{LSTM}{LSTM}{Long Short-Term Memory Network}
\newacronym{GRU}{GRU}{Gated Recurrent Unit}

\newacronym{IA}{IA}{Intra-Attention Network}
\newacronym{MHA}{MHA}{Multi-Headed Attention Network}
\newacronym{GNTP}{GNTP}{Greedy NTP}
\newacronym{CTP}{CTP}{Conditional Theorem Prover}

\newacronym{RNN}{RNN}{Recurrent Neural Network}
\newacronym{NMN}{NMN}{Neural Module Network}

\newacronym{CLUTRR}{CLUTRR}{Compositional Language Understanding and Text-based Relational Reasoning}

%% file: macros.tex
\newcommand{\eg}{\emph{e.g.}}
\newcommand{\ie}{\emph{i.e.}}

\newcommand{\kb}{\mathcal{K}}

\newcommand{\state}{S}

\newcommand{\krnl}{K}

\renewcommand{\emptyset}{\varnothing}

\def\rels{\mathcal{R}}

\let\set\dom 
\newcommand{\lss}[1]{\mathbb{#1}}
\let\ls\textsc
\let\lst\ls

\newcommand{\xs}[1]{\bm{[} #1 \bm{]}}

\renewcommand\vec[1]{{\bm{#1}}}

\makeatletter
\newcommand*\bdot{\mathpalette\bdot@{.5}}
\newcommand*\bdot@[2]{\mathbin{\vcenter{\hbox{\scalebox{#2}{$\m@th#1\bullet$}}}}}
\makeatother

\def\params{{\vec{\theta}}}

\newcommand{\rel}[1]{\verb~#1~}
\newcommand{\const}[1]{\textsc{#1}}

\newcommand{\var}[1]{\textsc{#1}}
\def\lif{\ \text{:--}\ }

\let\land\wedge

\let\subs\psi

\newcommand{\pred}[1]{\text{$\verb~#1~$}}

\def \Real {\mathbb{R}}

\let\success\rho

\let\todonote\todo

\colorlet{fixme}{red!85!black}
\colorlet{fixme-bright}{fixme!25}

\colorlet{todo}{orange!85!black}
\colorlet{todo-bright}{todo!25}

\colorlet{review}{nice-yellow!50!nice-orange}
\colorlet{review-bright}{review!25}

\colorlet{maybe}{nice-yellow}
\colorlet{maybe-bright}{maybe!25}

\colorlet{toref}{purple!70!black}
\colorlet{toref-bright}{toref!25}

\colorlet{info}{green!50!black}
\colorlet{info-bright}{info!25}

\colorlet{cut}{black!60}
\colorlet{cut-bright}{cut!25}

\colorlet{extend}{blue!85!black}
\colorlet{extend-bright}{extend!25}

\colorlet{discuss}{nice-red!50!black}
\colorlet{discuss-bright}{discuss!25}

\renewcommand{\todo}[1]{\todonote[linecolor=todo, backgroundcolor=todo-bright, bordercolor=todo]{\thesection{} {\bf\color{todo}TODO} #1}{}}

\newcommand{\info}[1]{}
\newcommand{\hlinfo}[2]{}

\newcommand{\maybe}[1]{}
\newcommand{\hlmaybe}[2]{}

\renewcommand{\const}[1]{{\color{black}\textsc{#1}}}

\renewcommand{\var}[1]{{\color{black}\textsc{#1}}}

\newcommand{\module}[1]{{\verb~#1~}}

%% file: tables/rules.tex
%
\begin{figure}[t!]
\begin{center}
\resizebox{0.9\columnwidth}{!}{
\begin{tabular}{rl}
\toprule
\multicolumn{2}{c}{\bf CLUTRR -- Sample of Learned Rules} \\
\midrule
$\rel{child}(\var{X}, \var{Y})$ & $\Leftarrow \rel{child}(\var{X}, \var{Z}), \rel{sibling}(\var{Z}, \var{Y})$ \\
$\rel{child}(\var{X}, \var{Y})$ & $\Leftarrow \rel{SO}(\var{X}, \var{Z}), \rel{child}(\var{Z}, \var{Y})$ \\
\midrule
$\rel{grand}(\var{X}, \var{Y})$ & $\Leftarrow \rel{child}(\var{X}, \var{Z}), \rel{child}(\var{Z}, \var{Y})$ \\
$\rel{grand}(\var{X}, \var{Y})$ & $\Leftarrow \rel{grand}(\var{X}, \var{Z}), \rel{sibling}(\var{Z}, \var{Y})$ \\
$\rel{grand}(\var{X}, \var{Y})$ & $\Leftarrow \rel{SO}(\var{X}, \var{Z}), \rel{grand}(\var{Z}, \var{Y})$ \\
\midrule
$\rel{in-law}(\var{X}, \var{Y})$ & $\Leftarrow \rel{child}(\var{X}, \var{Z}), \rel{SO}(\var{Z}, \var{Y})$ \\
$\rel{in-law}(\var{X}, \var{Y})$ & $\Leftarrow \rel{sibling-in-law}(\var{X}, \var{Z}), \rel{child}(\var{Z}, \var{Y})$ \\
\midrule
$\rel{sibling}(\var{X}, \var{Y})$ & $\Leftarrow \rel{sibling}(\var{X}, \var{Z}), \rel{sibling}(\var{Z}, \var{Y})$ \\
$\rel{sibling}(\var{X}, \var{Y})$ & $\Leftarrow \rel{child}(\var{X}, \var{Z}), \rel{uncle}(\var{Y}, \var{Z})$ \\
$\rel{sibling}(\var{X}, \var{Y})$ & $\Leftarrow \rel{child}(\var{X}, \var{Z}), \rel{child}(\var{Y}, \var{Z})$ \\
\bottomrule
\end{tabular}
}
\end{center}
\caption{Rules learned on \gls{CLUTRR} by \glspl{CTP} -- symbols were obtained by decoding the goal reformulations with the nearest predicate in embedding space.} \label{tab:clutrr-rules}
\end{figure}
%

%% file: figures/089907f8.tex
\begin{figure*}[t]
    \begin{center}
        \begin{minipage}{0.495\textwidth}
            \begin{center}
                \begin{minipage}{\textwidth}
                    \begin{center}
                        \includegraphics[width=0.95\textwidth]{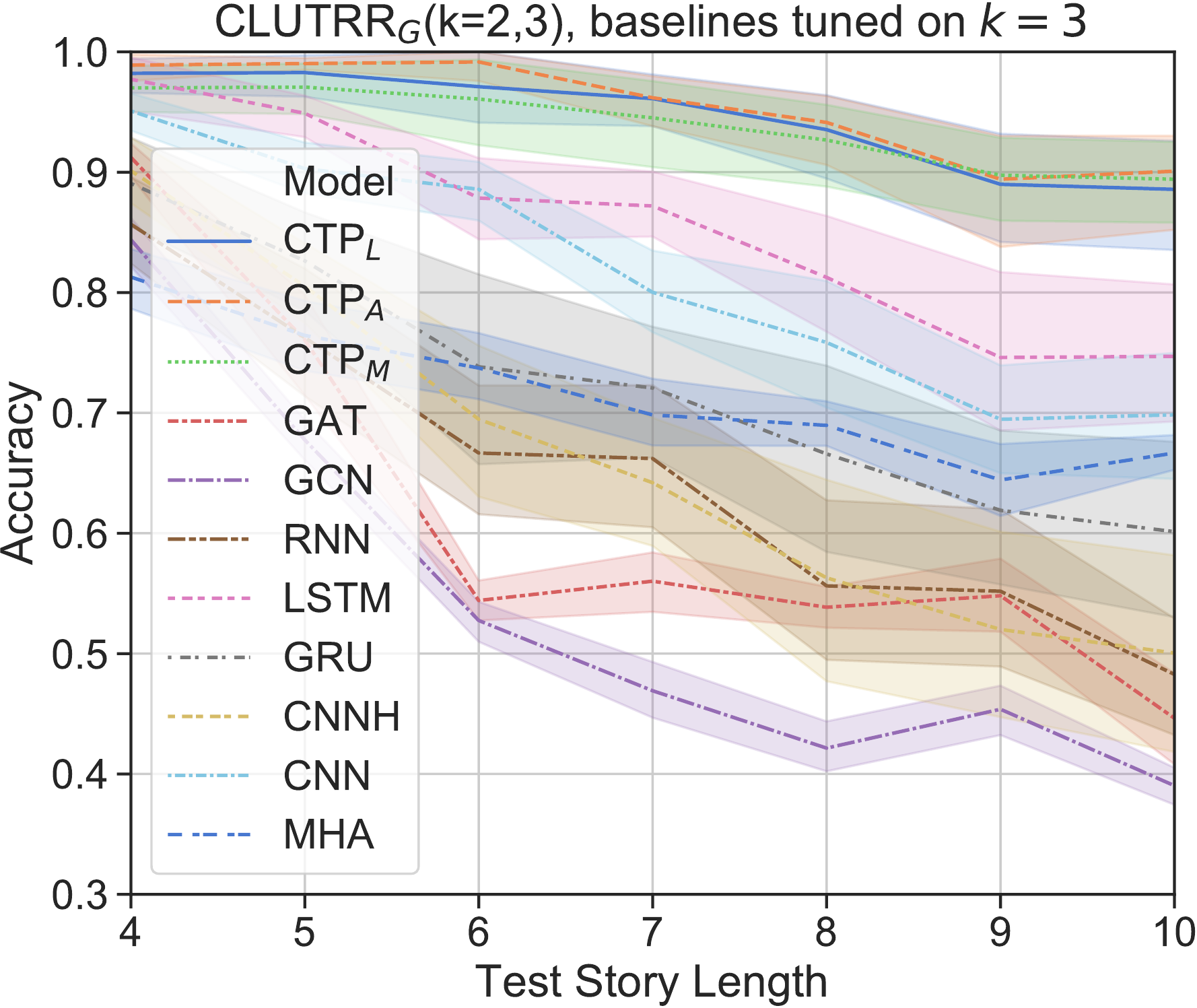}
                    \end{center}
                \end{minipage}
                \begin{minipage}{1.00\textwidth}
                    \begin{center}
                        \include{tables/089907f8_3}

                    \end{center}
                \end{minipage}
            \end{center}
        \end{minipage}
        \begin{minipage}{0.495\textwidth}
            \begin{center}
                \begin{minipage}{\textwidth}
                    \begin{center}
                        \includegraphics[width=0.95\textwidth]{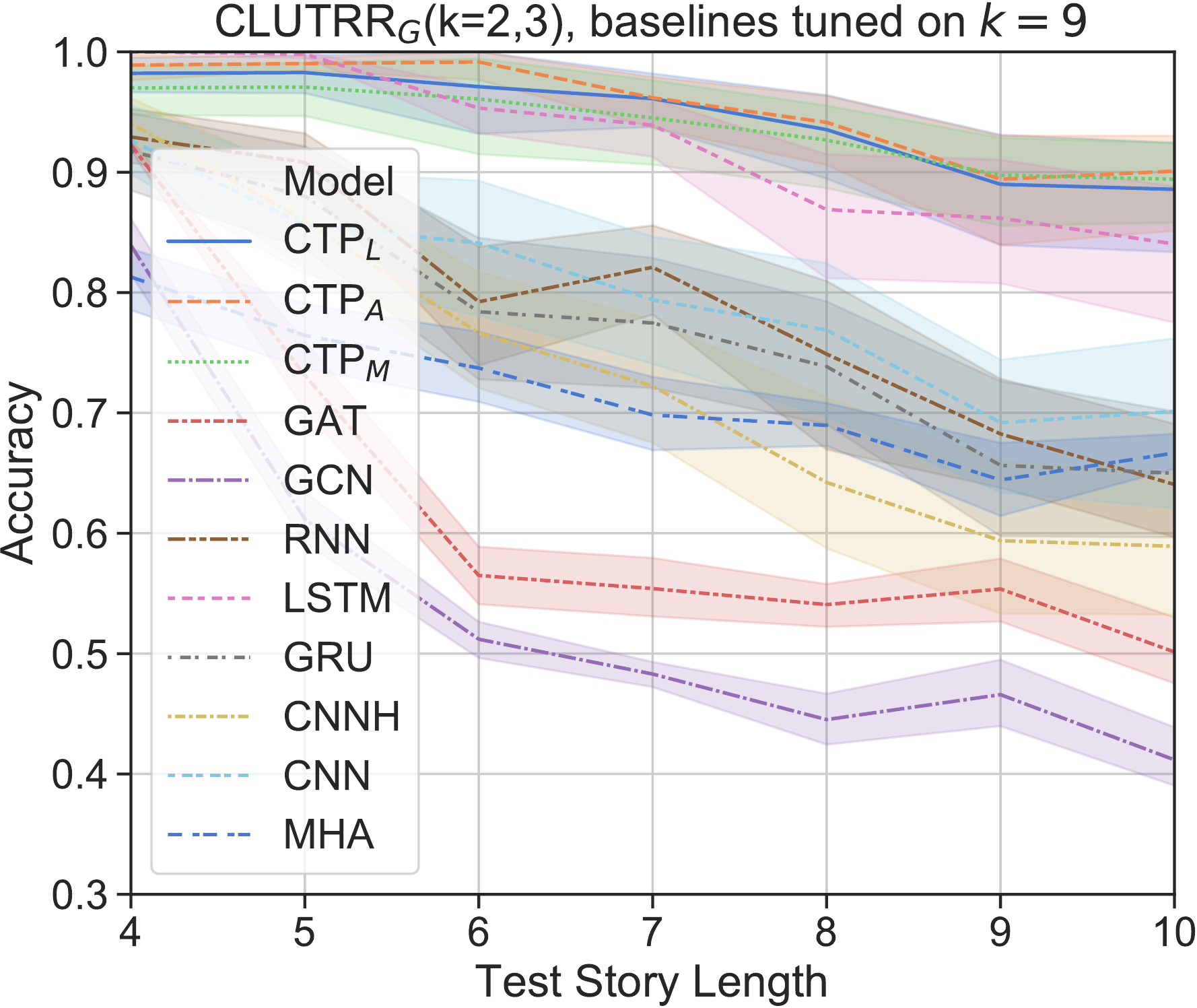}
                    \end{center}
                \end{minipage}
                \begin{minipage}{1.00\textwidth}
                    \begin{center}
                        \include{tables/089907f8_9}

                    \end{center}
                \end{minipage}
            \end{center}
        \end{minipage}
    \end{center}
\caption{Results on the \glsentryshort{CLUTRR} dataset after training on stories of lengths $\{ 2, 3 \}$ and evaluating on stories of length $\{ 4, 5, \ldots, 10 \}$ -- hyperparameters were fine-tuned on either short stories (left) and long stories (right).
Significance testing was assessed via a unequal variances $t$-test in comparison with CTP$_{L}$: $\blacktriangledown$ (resp. $\triangledown$) represents a $p$-value lower than $10^{-4}$ (resp. $10^{-2}$).
} \label{fig:089907f8}
\end{figure*}

%% file: tables/089907f8_3.tex
\begin{center}
\resizebox{\columnwidth}{!}{
\begin{tabular}{rlllllll}
\toprule
 & \multicolumn{1}{c}{\bf 4 Hops} & \multicolumn{1}{c}{\bf 5 Hops} & \multicolumn{1}{c}{\bf 6 Hops} & \multicolumn{1}{c}{\bf 7 Hops} & \multicolumn{1}{c}{\bf 8 Hops} & \multicolumn{1}{c}{\bf 9 Hops} & \multicolumn{1}{c}{\bf 10 Hops} \\
\midrule
 CTP$_{L}$  & $.98 \pm .02$  & $.98 \pm .03$  & $.97 \pm .05$  & $.96 \pm .04$  & $.94 \pm .05$  & $.89 \pm .07$  & $.89 \pm .07$  \\
 CTP$_{A}$  & $\mathbf{.99 \pm .02}$  & $\mathbf{.99 \pm .01}$  & $\mathbf{.99 \pm .02}$  & $\mathbf{.96 \pm .04}$  & $\mathbf{.94 \pm .05}$  & $.89 \pm .08$  & $\mathbf{.90 \pm .07}$  \\
 CTP$_{M}$  & $.97 \pm .03$  & $.97 \pm .03$  & $.96 \pm .06$  & $.95 \pm .06$  & $.93 \pm .06$  & $\mathbf{.90 \pm .06}$  & $.89 \pm .06$  \\
\midrule
 GNTP  & $.49 \pm .18$ $\triangledown$  & $.45 \pm .21$ $\triangledown$  & $.38 \pm .23$ $\triangledown$  & $.37 \pm .21$ $\triangledown$  & $.32 \pm .20$ $\triangledown$  & $.31 \pm .19$ $\triangledown$  & $.31 \pm .22$ $\triangledown$  \\
 \midrule
 GAT  & $.91 \pm .02$ $\blacktriangledown$  & $.76 \pm .06$ $\blacktriangledown$  & $.54 \pm .03$ $\blacktriangledown$  & $.56 \pm .04$ $\blacktriangledown$  & $.54 \pm .03$ $\blacktriangledown$  & $.55 \pm .05$ $\blacktriangledown$  & $.45 \pm .06$ $\blacktriangledown$  \\
 GCN  & $.84 \pm .03$ $\blacktriangledown$  & $.68 \pm .02$ $\blacktriangledown$  & $.53 \pm .03$ $\blacktriangledown$  & $.47 \pm .04$ $\blacktriangledown$  & $.42 \pm .03$ $\blacktriangledown$  & $.45 \pm .03$ $\blacktriangledown$  & $.39 \pm .02$ $\blacktriangledown$  \\
 \midrule
 RNN  & $.86 \pm .06$ $\triangledown$  & $.76 \pm .08$ $\blacktriangledown$  & $.67 \pm .08$ $\blacktriangledown$  & $.66 \pm .08$ $\blacktriangledown$  & $.56 \pm .10$ $\blacktriangledown$  & $.55 \pm .10$ $\blacktriangledown$  & $.48 \pm .07$ $\blacktriangledown$  \\
 LSTM  & $.98 \pm .04$  & $.95 \pm .03$  & $.88 \pm .05$ $\triangledown$  & $.87 \pm .04$ $\triangledown$  & $.81 \pm .07$ $\triangledown$  & $.75 \pm .10$ $\triangledown$  & $.75 \pm .09$ $\triangledown$  \\
 GRU  & $.89 \pm .05$ $\triangledown$  & $.83 \pm .06$ $\blacktriangledown$  & $.74 \pm .12$ $\triangledown$  & $.72 \pm .09$ $\blacktriangledown$  & $.67 \pm .12$ $\triangledown$  & $.62 \pm .10$ $\blacktriangledown$  & $.60 \pm .12$ $\blacktriangledown$  \\
 \midrule
 CNNH  & $.90 \pm .04$ $\triangledown$  & $.81 \pm .05$ $\blacktriangledown$  & $.69 \pm .10$ $\blacktriangledown$  & $.64 \pm .08$ $\blacktriangledown$  & $.56 \pm .13$ $\blacktriangledown$  & $.52 \pm .12$ $\blacktriangledown$  & $.50 \pm .12$ $\blacktriangledown$  \\
 CNN  & $.95 \pm .02$  & $.90 \pm .03$ $\blacktriangledown$  & $.89 \pm .04$ $\triangledown$  & $.80 \pm .05$ $\blacktriangledown$  & $.76 \pm .08$ $\triangledown$  & $.69 \pm .07$ $\blacktriangledown$  & $.70 \pm .08$ $\triangledown$  \\
 MHA  & $.81 \pm .04$ $\blacktriangledown$  & $.76 \pm .04$ $\blacktriangledown$  & $.74 \pm .05$ $\blacktriangledown$  & $.70 \pm .04$ $\blacktriangledown$  & $.69 \pm .03$ $\blacktriangledown$  & $.64 \pm .05$ $\blacktriangledown$  & $.67 \pm .02$ $\blacktriangledown$  \\
\bottomrule
\end{tabular}
}
\end{center}

%% file: tables/089907f8_9.tex
\begin{center}
\resizebox{\columnwidth}{!}{
\begin{tabular}{rlllllll}
\toprule
 & \multicolumn{1}{c}{\bf 4 Hops} & \multicolumn{1}{c}{\bf 5 Hops} & \multicolumn{1}{c}{\bf 6 Hops} & \multicolumn{1}{c}{\bf 7 Hops} & \multicolumn{1}{c}{\bf 8 Hops} & \multicolumn{1}{c}{\bf 9 Hops} & \multicolumn{1}{c}{\bf 10 Hops} \\
\midrule
 CTP$_{L}$  & $.98 \pm .02$  & $.98 \pm .03$  & $.97 \pm .05$  & $.96 \pm .04$  & $.94 \pm .05$  & $.89 \pm .07$  & $.89 \pm .07$  \\
 CTP$_{A}$  & $.99 \pm .02$  & $.99 \pm .01$  & $\mathbf{.99 \pm .02}$  & $\mathbf{.96 \pm .04}$  & $\mathbf{.94 \pm .05}$  & $.89 \pm .08$  & $\mathbf{.90 \pm .07}$  \\
 CTP$_{M}$  & $.97 \pm .03$  & $.97 \pm .03$  & $.96 \pm .06$  & $.95 \pm .06$  & $.93 \pm .06$  & $\mathbf{.90 \pm .06}$  & $.89 \pm .06$  \\
\midrule
 GNTP  & $.49 \pm .18$ $\triangledown$  & $.45 \pm .21$ $\triangledown$  & $.38 \pm .23$ $\triangledown$  & $.37 \pm .21$ $\triangledown$  & $.32 \pm .20$ $\triangledown$  & $.31 \pm .19$ $\triangledown$  & $.31 \pm .22$ $\triangledown$  \\
 \midrule
 GAT  & $.92 \pm .01$ $\blacktriangledown$  & $.73 \pm .04$ $\blacktriangledown$  & $.56 \pm .04$ $\blacktriangledown$  & $.55 \pm .04$ $\blacktriangledown$  & $.54 \pm .03$ $\blacktriangledown$  & $.55 \pm .04$ $\blacktriangledown$  & $.50 \pm .04$ $\blacktriangledown$  \\
 GCN  & $.84 \pm .04$ $\blacktriangledown$  & $.61 \pm .03$ $\blacktriangledown$  & $.51 \pm .02$ $\blacktriangledown$  & $.48 \pm .02$ $\blacktriangledown$  & $.45 \pm .03$ $\blacktriangledown$  & $.47 \pm .05$ $\blacktriangledown$  & $.41 \pm .04$ $\blacktriangledown$  \\
 \midrule
 RNN  & $.93 \pm .03$ $\triangledown$  & $.91 \pm .03$ $\triangledown$  & $.79 \pm .08$ $\blacktriangledown$  & $.82 \pm .06$ $\blacktriangledown$  & $.75 \pm .11$ $\triangledown$  & $.68 \pm .07$ $\blacktriangledown$  & $.64 \pm .07$ $\blacktriangledown$  \\
 LSTM  & $\mathbf{1.0 \pm .00}$  & $\mathbf{1.0 \pm .00}$  & $.95 \pm .03$  & $.94 \pm .04$  & $.87 \pm .08$  & $.86 \pm .08$  & $.84 \pm .09$  \\
 GRU  & $.92 \pm .05$ $\triangledown$  & $.88 \pm .06$ $\triangledown$  & $.78 \pm .09$ $\triangledown$  & $.77 \pm .09$ $\triangledown$  & $.74 \pm .08$ $\blacktriangledown$  & $.66 \pm .10$ $\blacktriangledown$  & $.65 \pm .08$ $\blacktriangledown$  \\
 \midrule
 CNNH  & $.94 \pm .03$ $\triangledown$  & $.86 \pm .06$ $\triangledown$  & $.77 \pm .08$ $\blacktriangledown$  & $.72 \pm .08$ $\blacktriangledown$  & $.64 \pm .09$ $\blacktriangledown$  & $.59 \pm .10$ $\blacktriangledown$  & $.59 \pm .09$ $\blacktriangledown$  \\
 CNN  & $.93 \pm .04$ $\triangledown$  & $.86 \pm .07$ $\triangledown$  & $.84 \pm .09$ $\triangledown$  & $.79 \pm .08$ $\triangledown$  & $.77 \pm .10$ $\triangledown$  & $.69 \pm .09$ $\triangledown$  & $.70 \pm .11$ $\triangledown$  \\
 MHA  & $.81 \pm .04$ $\blacktriangledown$  & $.76 \pm .04$ $\blacktriangledown$  & $.74 \pm .05$ $\blacktriangledown$  & $.70 \pm .04$ $\blacktriangledown$  & $.69 \pm .03$ $\blacktriangledown$  & $.64 \pm .05$ $\blacktriangledown$  & $.67 \pm .02$ $\blacktriangledown$  \\
\bottomrule
\end{tabular}
}
\end{center}

%% file: figures/db9b8f04.tex
\begin{figure*}[t]
\begin{center}
    \begin{minipage}{0.495\textwidth}
        \begin{center}
            \begin{minipage}{\textwidth}
                \begin{center}
                    \includegraphics[width=0.95\textwidth]{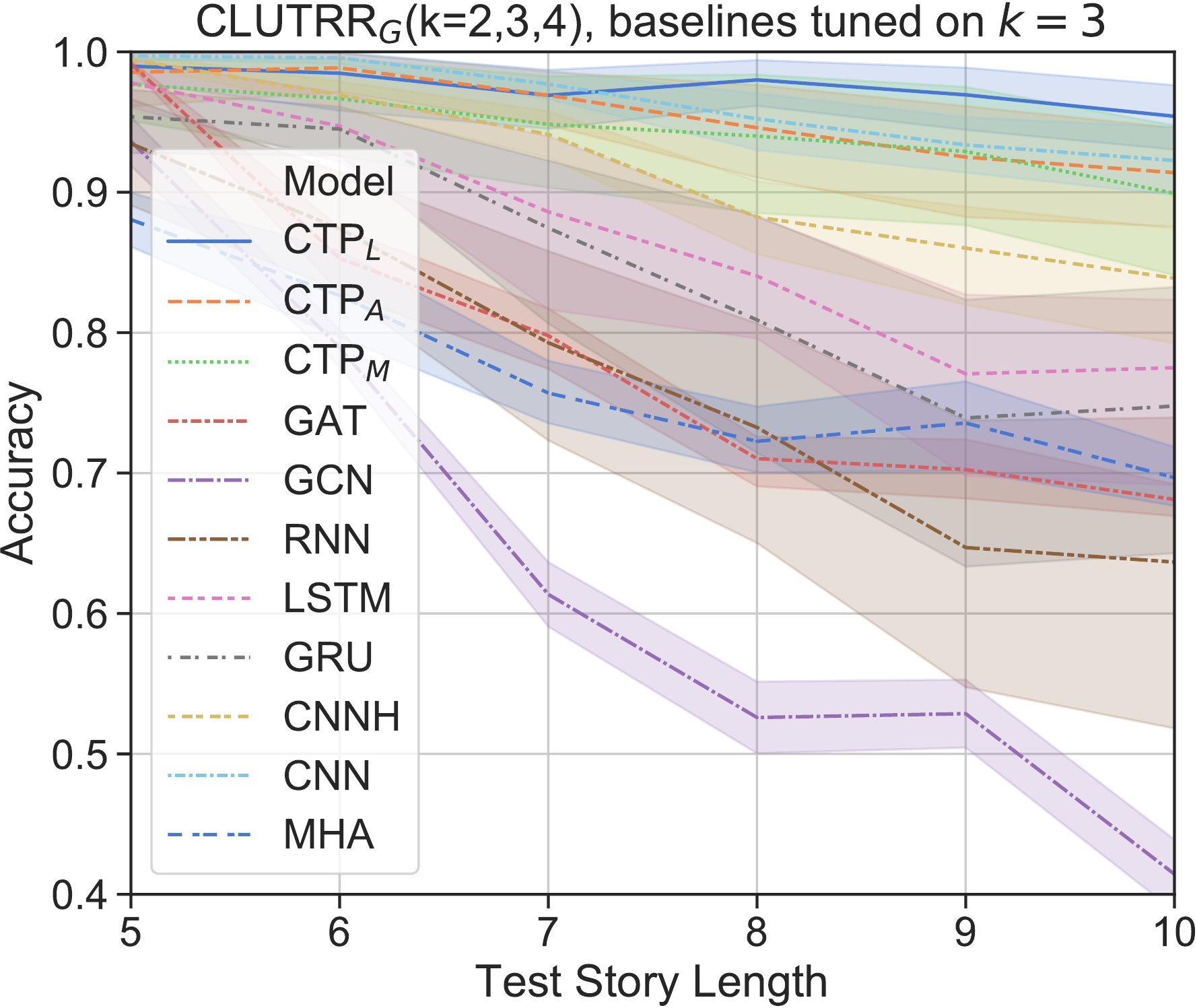}
                \end{center}
            \end{minipage}
            \begin{minipage}{\textwidth}
                \begin{center}
                    \include{tables/db9b8f04_3}

                \end{center}
            \end{minipage}
        \end{center}
    \end{minipage}
    \begin{minipage}{0.495\textwidth}
        \begin{center}
            \begin{minipage}{\textwidth}
                \begin{center}
                    \includegraphics[width=0.95\textwidth]{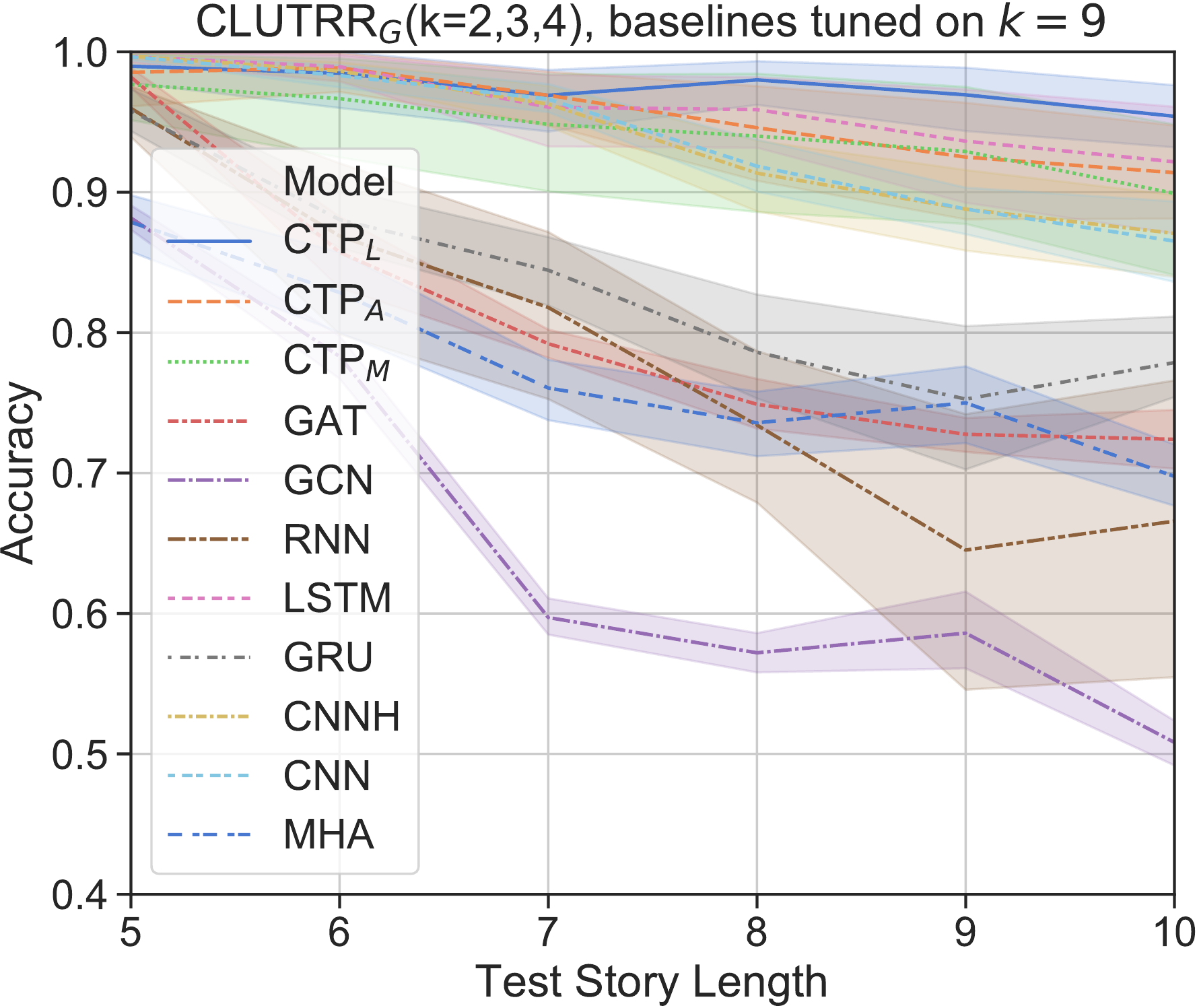}
                \end{center}
            \end{minipage}
            \begin{minipage}{\textwidth}
                \begin{center}
                    \include{tables/db9b8f04_9}

                \end{center}
            \end{minipage}
        \end{center}
    \end{minipage}
\end{center}
\caption{Results on the \glsentryshort{CLUTRR} dataset after training on stories of lengths $\{ 2, 3, 4 \}$ and evaluating on stories of length $\{ 5, \ldots, 10 \}$ -- hyperparameters were fine-tuned on either short stories (left) and long stories (right). Significance testing was assessed via a unequal variances $t$-test in comparison with CTP$_{L}$: $\blacktriangledown$ (resp. $\triangledown$) represents a $p$-value lower than $10^{-4}$ (resp. $10^{-2}$).
} \label{fig:db9b8f04}
\end{figure*}

%% file: tables/db9b8f04_3.tex
\begin{center}
\resizebox{\columnwidth}{!}{
\begin{tabular}{rllllll}
\toprule
 & \multicolumn{1}{c}{\bf 5 Hops} & \multicolumn{1}{c}{\bf 6 Hops} & \multicolumn{1}{c}{\bf 7 Hops} & \multicolumn{1}{c}{\bf 8 Hops} & \multicolumn{1}{c}{\bf 9 Hops} & \multicolumn{1}{c}{\bf 10 Hops} \\
\midrule
 CTP$_{L}$  & $.99 \pm .02$  & $.98 \pm .04$  & $.97 \pm .04$  & $\mathbf{.98 \pm .03}$  & $\mathbf{.97 \pm .04}$  & $\mathbf{.95 \pm .04}$  \\
 CTP$_{A}$  & $.99 \pm .04$  & $.99 \pm .03$  & $.97 \pm .03$  & $.95 \pm .06$  & $.93 \pm .07$  & $.91 \pm .05$  \\
 CTP$_{M}$  & $.98 \pm .04$  & $.97 \pm .06$  & $.95 \pm .06$  & $.94 \pm .08$  & $.93 \pm .08$  & $.90 \pm .09$  \\
\midrule
 GNTP  & $.68 \pm .28$  & $.63 \pm .34$  & $.62 \pm .31$  & $.59 \pm .32$  & $.57 \pm .34$  & $.52 \pm .32$  \\
 \midrule
 GAT  & $.99 \pm .00$  & $.85 \pm .04$ $\blacktriangledown$  & $.80 \pm .03$ $\blacktriangledown$  & $.71 \pm .03$ $\blacktriangledown$  & $.70 \pm .03$ $\blacktriangledown$  & $.68 \pm .02$ $\blacktriangledown$  \\
 GCN  & $.94 \pm .03$ $\triangledown$  & $.79 \pm .02$ $\blacktriangledown$  & $.61 \pm .03$ $\blacktriangledown$  & $.53 \pm .04$ $\blacktriangledown$  & $.53 \pm .04$ $\blacktriangledown$  & $.41 \pm .04$ $\blacktriangledown$  \\
 \midrule
 RNN  & $.93 \pm .06$  & $.87 \pm .07$ $\triangledown$  & $.79 \pm .11$ $\triangledown$  & $.73 \pm .12$ $\triangledown$  & $.65 \pm .16$ $\triangledown$  & $.64 \pm .16$ $\triangledown$  \\
 LSTM  & $.98 \pm .03$  & $.95 \pm .04$  & $.89 \pm .10$  & $.84 \pm .07$ $\triangledown$  & $.77 \pm .11$ $\triangledown$  & $.78 \pm .11$ $\triangledown$  \\
 GRU  & $.95 \pm .04$  & $.94 \pm .03$  & $.87 \pm .08$  & $.81 \pm .13$ $\triangledown$  & $.74 \pm .15$ $\triangledown$  & $.75 \pm .15$ $\triangledown$  \\
 \midrule
 CNNH  & $.99 \pm .01$  & $.97 \pm .02$  & $.94 \pm .03$  & $.88 \pm .04$ $\blacktriangledown$  & $.86 \pm .05$ $\triangledown$  & $.84 \pm .06$ $\triangledown$  \\
 CNN  & $\mathbf{1.0 \pm .00}$  & $\mathbf{1.0 \pm .01}$  & $\mathbf{.98 \pm .01}$  & $.95 \pm .03$  & $.93 \pm .03$  & $.92 \pm .04$  \\
 MHA  & $.88 \pm .03$ $\blacktriangledown$  & $.83 \pm .05$ $\blacktriangledown$  & $.76 \pm .04$ $\blacktriangledown$  & $.72 \pm .04$ $\blacktriangledown$  & $.74 \pm .05$ $\blacktriangledown$  & $.70 \pm .03$ $\blacktriangledown$  \\
\bottomrule
\end{tabular}
}
\end{center}

%% file: tables/db9b8f04_9.tex
\begin{center}
\resizebox{\columnwidth}{!}{
\begin{tabular}{rllllll}
\toprule
 & \multicolumn{1}{c}{\bf 5 Hops} & \multicolumn{1}{c}{\bf 6 Hops} & \multicolumn{1}{c}{\bf 7 Hops} & \multicolumn{1}{c}{\bf 8 Hops} & \multicolumn{1}{c}{\bf 9 Hops} & \multicolumn{1}{c}{\bf 10 Hops} \\
\midrule
 CTP$_{L}$  & $.99 \pm .02$  & $.98 \pm .04$  & $.97 \pm .04$  & $\mathbf{.98 \pm .03}$  & $\mathbf{.97 \pm .04}$  & $\mathbf{.95 \pm .04}$  \\
 CTP$_{A}$  & $.99 \pm .04$  & $.99 \pm .03$  & $\mathbf{.97 \pm .03}$  & $.95 \pm .06$  & $.93 \pm .07$  & $.91 \pm .05$  \\
 CTP$_{M}$  & $.98 \pm .04$  & $.97 \pm .06$  & $.95 \pm .06$  & $.94 \pm .08$  & $.93 \pm .08$  & $.90 \pm .09$  \\
\midrule
 GNTP  & $.68 \pm .28$  & $.63 \pm .34$  & $.62 \pm .31$  & $.59 \pm .32$  & $.57 \pm .34$  & $.52 \pm .32$  \\
 \midrule
 GAT  & $.98 \pm .01$  & $.86 \pm .04$ $\blacktriangledown$  & $.79 \pm .02$ $\blacktriangledown$  & $.75 \pm .03$ $\blacktriangledown$  & $.73 \pm .02$ $\blacktriangledown$  & $.72 \pm .03$ $\blacktriangledown$  \\
 GCN  & $.88 \pm .01$ $\blacktriangledown$  & $.78 \pm .02$ $\blacktriangledown$  & $.60 \pm .02$ $\blacktriangledown$  & $.57 \pm .02$ $\blacktriangledown$  & $.59 \pm .04$ $\blacktriangledown$  & $.51 \pm .02$ $\blacktriangledown$  \\
 \midrule
 RNN  & $.96 \pm .03$  & $.87 \pm .09$ $\triangledown$  & $.82 \pm .09$ $\triangledown$  & $.73 \pm .09$ $\blacktriangledown$  & $.65 \pm .15$ $\triangledown$  & $.67 \pm .16$ $\triangledown$  \\
 LSTM  & $1.0 \pm .01$  & $\mathbf{.99 \pm .02}$  & $.96 \pm .04$  & $.96 \pm .04$  & $.94 \pm .06$  & $.92 \pm .07$  \\
 GRU  & $.96 \pm .02$ $\triangledown$  & $.88 \pm .03$ $\blacktriangledown$  & $.84 \pm .04$ $\blacktriangledown$  & $.79 \pm .06$ $\blacktriangledown$  & $.75 \pm .08$ $\blacktriangledown$  & $.78 \pm .04$ $\blacktriangledown$  \\
 \midrule
 CNNH  & $\mathbf{1.0 \pm .00}$  & $.99 \pm .01$  & $.96 \pm .02$  & $.91 \pm .04$ $\triangledown$  & $.89 \pm .04$ $\triangledown$  & $.87 \pm .04$ $\triangledown$  \\
 CNN  & $1.0 \pm .00$  & $.98 \pm .01$  & $.97 \pm .02$  & $.92 \pm .03$ $\triangledown$  & $.89 \pm .03$ $\blacktriangledown$  & $.87 \pm .04$ $\triangledown$  \\
 MHA  & $.88 \pm .03$ $\blacktriangledown$  & $.83 \pm .05$ $\blacktriangledown$  & $.76 \pm .03$ $\blacktriangledown$  & $.74 \pm .04$ $\blacktriangledown$  & $.75 \pm .04$ $\blacktriangledown$  & $.70 \pm .03$ $\blacktriangledown$  \\
\bottomrule
\end{tabular}
}
\end{center}

%% file: tables/small_rules.tex
\begin{table*}[t]
\caption{
Comparison of \glspl{CTP}, with \glspl{GNTP}~\citep{DBLP:conf/aaai/MinerviniBR0G20}, NeuralLP~\citep{DBLP:conf/nips/YangYC17} and MINERVA~\citep{DBLP:conf/iclr/DasDZVDKSM18} (from \citet{DBLP:conf/aaai/MinerviniBR0G20}) on benchmark datasets: hyperparameters were selected based on the validation MRR, and we report the mean and standard deviation over 10 random seeds.
}
\label{tab:benchmark}
\begin{center}
\resizebox{\textwidth}{!}{
\setlength{\tabcolsep}{0.2em}
\begin{tabular}{cclcccccl}
\toprule
\multicolumn{2}{c}{\multirow{4}{*}{\bf Datasets}} & \multirow{4}{*}{\bf Metrics} & \multicolumn{5}{c}{\bf Models} & \multirow{4}{*}{\bf \qquad Learned Rules} \\
& & & \multirow{2}{*}{\bf CTP} & \multicolumn{2}{c}{\bf GNTP} & \multirow{2}{*}{\bf NeuralLP} & \multirow{2}{*}{\bf MINERVA} \\
\cmidrule{5-6}
& & & & {\bf Standard} & {\bf Attention} \\
\midrule
\multirow{3}{*}{\bf Countries} & $S1$ & \multirow{3}{*}{AUC-PR} & {\bf 100.0 $\pm$ 0.00} & 99.98 $\pm$ 0.05 & {\bf 100.0 $\pm$ 0.00} & {\bf 100.0 $\pm$ 0.0} & {\bf 100.0 $\pm$ 0.0} & \rel{locIn}(\var{X},\var{Y}) \lif \rel{locIn}(\var{X},\var{Z}), \rel{locIn}(\var{Z},\var{Y}) \\
& $S2$ & & 91.81 $\pm$ 1.07 & 90.82 $\pm$ 0.88 & \textbf{93.48 $\pm$ 3.29} & 75.1 $\pm$ 0.3 & 92.36 $\pm$ 2.41 & \rel{neighOf}(\var{X},\var{Y}) \lif \rel{neighOf}(\var{X},\var{Z}), \rel{locIn}(\var{Z},\var{Y}) \\
& $S3$ & & 94.78 $\pm$ 0.00 & 87.70 $\pm$ 4.79 & 91.27 $\pm$ 4.02 & 92.20 $\pm$ 0.2 & {\bf 95.10 $\pm$ 1.20} & \rel{neighOf}(\var{X},\var{Y}) \lif \rel{neighOf}(\var{Y},\var{X}) \\
\midrule
\multirow{4}{*}{\bf Kinship}
%
%
%
%
%
& & MRR & {\bf 0.764 $\pm$ 0.00} & 0.719 & 0.759 & 0.619 & 0.720 & \rel{term0}(\var{X}, \var{Y}) \lif \rel{term22}(\var{Y}, \var{X})\\
%
& & Hits@1 & {\bf 0.646 $\pm$ 0.01} & 0.586 & 0.642 & 0.475 & 0.605 & \rel{term4}(\var{X}, \var{Y}) \lif \rel{term4}(\var{Y}, \var{X}) \\
%
& & Hits@3 & {\bf 0.859 $\pm$ 0.01} & 0.815 & 0.850 & 0.707 & 0.812 & \rel{term20}(\var{X},\var{Y}) \lif \rel{term24}(\var{X}, \var{Z}), \rel{term6}(\var{Z}, \var{Y}) \\
%
& & Hits@10 & 0.958 $\pm$ 0.00 & 0.958 & {\bf 0.959} & 0.912 & 0.924 & \rel{term2}(\var{X},\var{Y}) \lif \rel{term4}(\var{X}, \var{Z}), \rel{term7}(\var{Z}, \var{Y})\\
\midrule
%
%
\multirow{4}{*}{\bf Nations}
%
%
& & MRR & {\bf 0.709 $\pm$ 0.03} & 0.658 & 0.645 & --- & --- & \rel{tourism3}(\var{X}, \var{Y}) \lif \rel{eemigrants}(\var{Y}, \var{X}) \\
%
& & Hits@1 & {\bf 0.562 $\pm$ 0.05} & 0.493 & 0.490 & --- & --- & \rel{independence}(\var{X},\var{Y}) \lif \rel{commonbloc0}(\var{Y},\var{X}) \\
%
& & Hits@3 & {\bf 0.813 $\pm$ 0.03} & 0.781 & 0.736 & --- & --- & \rel{relngo}(\var{X},\var{Y}) \lif \rel{timesinceally}(\var{Y},\var{X}) \\
%
& & Hits@10 & {\bf 0.995 $\pm$ 0.00} & 0.985 & 0.975 & --- & --- & \rel{relstudents}(\var{X}, \var{Y}) \lif \rel{relexportbooks}(\var{X}, \var{Y}) \\
\midrule
\multirow{4}{*}{\bf UMLS}
%
%
& & MRR & 0.852 $\pm$ 0.01 & 0.841 & {\bf 0.857} & 0.778 & 0.825 & \rel{isa}(\var{X},\var{Y}) \lif \rel{isa}(\var{X},\var{Z}), \rel{isa}(\var{Z},\var{Y}) \\
& & Hits@1 & 0.752 $\pm$ 0.01 & 0.732 & {\bf 0.761} & 0.643 & 0.728 & \rel{resultOf}(\var{X},\var{Y}) \lif \rel{resultOf}(\var{X},\var{Z}), \rel{resultOf}(\var{Z},\var{Y}) \\
& & Hits@3 & {\bf 0.947} $\pm$ 0.01 & 0.941 & {\bf 0.947} & 0.869 & 0.900 & \rel{treats}(\var{X}, \var{Y}) \lif \rel{prevents}(\var{X}, \var{Z}), \rel{resultOf}(\var{Z}, \var{Y}) \\
& & Hits@10 & 0.984 $\pm$ 0.00 & {\bf 0.986} & 0.983 & 0.962 & 0.968 & \rel{uses}(\var{X},\var{Y}) \lif \rel{produces}(\var{X},\var{Y}) \\
\bottomrule
\end{tabular}
}
\end{center}
\end{table*}